\documentclass[10pt,twocolumn,letterpaper]{article}
\usepackage{iccv}

\usepackage{url}
\usepackage[pagebackref=true,breaklinks=true,letterpaper=true,colorlinks,bookmarks=false]{hyperref}
\usepackage{morefloats}
\usepackage{graphicx}
\usepackage{pgfplots}
\usetikzlibrary{shapes,shadows,positioning,calc,patterns}
\usepackage{rotating}
\usepackage{multirow}
\usepackage{picinpar}
\usepackage{changepage}
\usepackage{wrapfig}
\usepackage[lined,boxed,commentsnumbered,ruled]{algorithm2e}
\usepackage{epstopdf}
\usepackage{color}

\input{Definitions}

\iccvfinalcopy 

\def\iccvPaperID{1159} 

\ificcvfinal\fi

\begin{document}

\title{Probabilistic Label Relation Graphs with Ising Models}

\author{
Nan Ding\\
Google Inc.\\
{\tt\small dingnan@google.com}
\and
Jia Deng\\
        University of Michigan\\
       {\tt\small jiadeng@umich.edu}
\and
\and
Kevin P. Murphy\\
        Google Inc.\\
       {\tt\small kpmurphy@google.com}
\and
Hartmut Neven\\
        Google Inc.\\
       {\tt\small neven@google.com}
}

\maketitle

\begin{abstract}
We consider classification problems in which the label space has structure. A common example is hierarchical label spaces, corresponding to the case where one label subsumes another (e.g., animal subsumes dog). But labels can also be mutually exclusive (e.g., dog vs cat) or unrelated (e.g., furry, carnivore). To jointly model hierarchy and exclusion relations, the notion of a HEX (hierarchy and exclusion) graph was introduced in \cite{DenDinJia14}. This combined a conditional random field (CRF) with a deep neural network (DNN), resulting in state of the art results when applied to visual object classification problems where the training labels were drawn from different levels of the ImageNet hierarchy (e.g., an image might be labeled with the basic level category "dog'', rather than the more specific label "husky''). In this paper, we extend the HEX model to allow for soft or probabilistic relations between labels, which is useful when there is uncertainty about the relationship between two labels (e.g., an antelope is "sort of'' furry, but not to the same degree as a grizzly bear). We call our new model pHEX, for probabilistic HEX. We show that the pHEX graph can be converted to an Ising model, which allows us to use existing off-the-shelf inference methods (in contrast to the HEX method, which needed specialized inference algorithms). Experimental results show significant improvements in a number of large-scale visual object classification tasks, outperforming the previous HEX model.
\end{abstract}


\section{Introduction}

Classification is a fundamental problem in machine learning and computer vision.
In this paper, we consider how to extend the standard approach to exploit structure in the label space.
For example, consider the problem of classifying images of animals.
The labels may be names of animal types (e.g., dog, puppy, cat),
or attribute labels (e.g., yellow, furry, has-stripes).
Many of these labels are not semantically independent of each other. For example, a puppy is also a dog,
which is a hierarchical or subsumption relation; an animal cannot be both a dog and a cat, an exclusive relation;
but an animal can be yellow and furry, which is a non-relation.

In \cite{DenDinJia14}, an approach called Hierarchy and EXclusion (HEX) graphs was proposed for compactly representing such constraints between the labels. 
In particular, a probabilistic graphical model with deterministic or hard constraints between the binary label nodes was
proposed. These hard constraints cut down the feasible set of labels from $2^n$ (where $n$ is the number of labels) to something much smaller, allowing for efficient exact inference.
For example, if all labels are mutually exclusive, the HEX graph is a clique, and there are only $n+1$ valid label configurations.
This graphical model can be combined with any standard
discriminative classifier (such as deep neural networks),
resulting in a conditional random field (CRF) model with label constraints.

In this paper, we extend the HEX model by allowing for ``soft'' relationships between the labels.
We call this the pHEX model.
The pHEX model has five main advantages compared
to the HEX model.
First, it is a more realistic model, since the relationship between most labels is ``soft''.
For example, a lion may be mostly yellow, but it could also be another color.
Second, the pHEX model is easier to train, since the likelihood function
is smoother.
Third, we show how to perform inference in the pHEX model  by converting it to an Ising model, and then using standard
off-the-shelf tools such as belief propagation, or the emerging quantum optimization technology \cite{KadNis98}.
This is in contrast to the HEX case, which needed a specialized (and rather complex) algorithm to perform inference.
Fourth, we show how to combine binary labels with $k$-ary labels, something that wasn't possible with the original HEX model.
Finally,  we show that  the pHEX model outperforms the HEX model on a variety of visual object classification tasks.


\eat{
In summary, the main contributions of this paper are as follows:
we propose a new  model called pHEX for representing soft or probabilistic relationships between labels;
we propose a way to perform efficient inference in such models by converting them to Ising models and then using standard tools;
and we show that the pHEX model outperforms the HEX model (which in turn was previously
shown to outperform standard multiclass and multilabel classification methods) on certain visual object classification tasks.
}

\eat{
A fundamental task of machine learning and pattern recognition is object classification, which assigns semantic labels from a pool of candidate labels to an object. In real applications, the candidate label pool may contain a large varieties of labels which follow a long tailed distribution. The labels may contain object labels like dog, puppy, cat; as well as attribute labels like: yellow, stripes, eating fish and so on. Many of those labels are not semantically independent of each other. For example, a puppy is also a dog, a hierarchy relation; an object cannot be both a dog and a car, an exclusive relation.

To build a desired classification model, it would be beneficial to consider these semantic relations among the candidate labels and be able to transfer knowledge between the labels. A natural choice of representing random variables which are dependent on each other would be the probabilistic graphical models \cite{WaiJor03}.  In \cite{DenDinJia14}, a unified label relation graph framework called Hierarchy and EXclusion (HEX) graphs have been proposed. In the HEX graph, three types of label relations: exclusion, overlapping, and subsumption are combined to a graphical model for multi-label classification. The label relations in the HEX graphs are absolute relations. This allows significant reduction of the size of the legal state space which plays a crucial role in designing efficient inference algorithms. 

On the other hand, absolute label relations are not sufficient to model all types of dependencies in reality. Some labels are related to each other but the relation is not deterministic. For example, a taxi may be mostly yellow, but it could also be in other color. In addition, some labels may be more related (or exclusive) than others. For example, a dog looks more like a cat than a car. To consider these uncertain and relative information, we need to quantify those label relations. 

In this paper, we study object classification by considering non-absolute (probabilistic) label relations and propose a new probabilistic HEX (pHEX) graph. We show that the pHEX graph is a generalization of the HEX graph, and is mathematically equivalent to an Ising model. The inference on a pHEX graph can be done using off-shelf approximate inference methods, such as loopy belief propagation. In a number of numerical experiments, we empirically demonstrate that the classification models based on pHEX graphs improve the results obtained by the ones based on HEX graphs. 

The remainder of the paper is structured as follows. Section \ref{sec:hex} reviews the HEX graphs. Section \ref{sec:phex} introduces the probabilistic HEX graph as an Ising model as well as the classification models based on the pHEX graphs. We then discusses the inference problem of pHEX graph in Section \ref{sec:inference} and the practical implementation in Section \ref{sec:implementation}. Section \ref{sec:experiment} shows the experimental results. The paper is concluded in Section \ref{sec:conclusion}. 
}

\section{Related work}

There has been a lot of prior work on exploiting structure in the label space; we only have space to mention a few key papers here.
Conditional random fields \cite{Lafferty01crf,Vishy06} and structural SVMs \cite{Tso05svmstruct} are often used in structured prediction problems. In addition, in transfer learning \cite{RohrbachSSGS10,rohrbach11cvpr}, zero-shot learning \cite{lampert2009,Palatucci09}, and attribute-based recognition \cite{AkataPHS13,CVPR13Yu,SharmanskaQL12}, consistency between visual predictions and semantic relations are often enforced. 

More closely related to this paper is work that exploits hierarchical structure (e.g., \cite{zweig2007,marszalek2007,wu-RSS-14,Ordonez2013}), exclusive relations \cite{chen2011multi}, or both of them \cite{Dalvi2015, MirRavXu2015}. 
Recently \cite{DenDinJia14} proposed the HEX graph approach,
which subsumes a lot of prior work by modeling hierarhical and exclusive relations using graphical models.
We discuss this in more detail in Section~\ref{sec:hex}, since it forms the foundation for the current paper.

\section{The HEX model}
\label{sec:hex}

In a nutshell, HEX graphs are probabilistic graphical models with directed and undirected edges over a number of binary variables. Each binary variable represents a label and takes value from $\cbr{-1,1}$. Each edge or no-edge between any two labels represents one of three label relations: exclusion, hierarchy and non-relation. The combination of all pairwise label relations allows the HEX graph to characterize the legal and illegal state space of labels, as we explain below.

\subsection{HEX relations}

The three types of label relations in the HEX graph are defined as follows:

\paragraph{Exclusion}
When two nodes are connected by an {\em undirected edge}, this is called an exclusive relation. It means that the two labels cannot be both equal to 1. For example, an animal cannot be both a \emph{cat} and a \emph{dog}. So \emph{cat} and \emph{dog} are mutually exclusive. The legal state space for exclusion is: 
\begin{equation}
S^e \defeq \cbr{(-1,-1), (-1,1), (1,-1)}.
\end{equation}
\eat{
\begin{figure}[!h]
  \centering
  \includegraphics[width=2in]{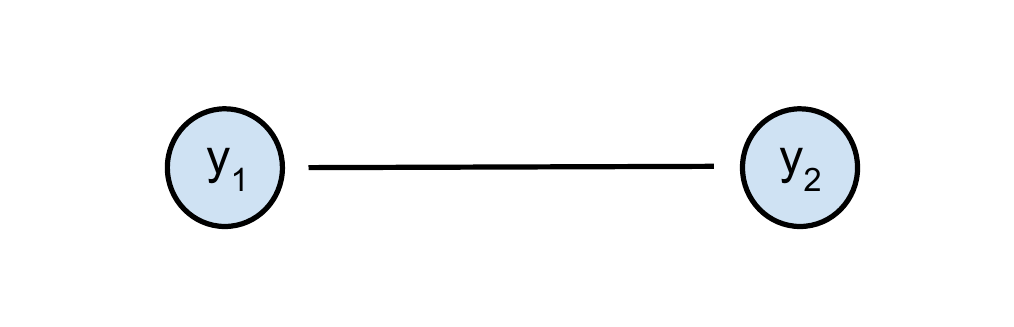}
  \caption{Exclusive relation in HEX graph: both $y_1$ and $y_2$ equal to 1 is prohibited.}
\label{fig:hex-exclusion}
\end{figure}
}

\paragraph{Hierarchy}
When two nodes are connected by a {\em directed edge} from $y_1$ to $y_2$, this is called a subsumption (hierarchical) relation. It means that if $y_2$ is 1 then $y_1$ must be 1 as well. For example, a \emph{puppy} is always a \emph{dog}. So \emph{dog} subsumes \emph{puppy}. The corresponding legal state space for subsumption is: 
\begin{equation}
S^h \defeq \cbr{(-1,-1),  (1,-1), (1,1)}.
\end{equation}

\eat{
\begin{figure}[!h]
  \centering
\includegraphics[width=2in]{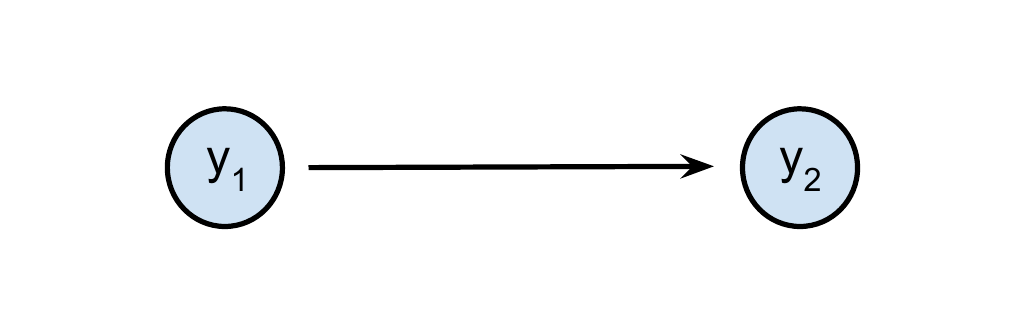}
  \caption{Subsumption (hierarchical) relation in HEX graph: if $y_2 = 1$, then $y_1$ has to be 1. }
\label{fig:hex-subsumption}
\end{figure}
}

\paragraph{No relation}

When two nodes are {\em not connected} by any edge, 
we say there is no relation between them.
This means that the two labels are independent of each other. For
example, \emph{carnivore} and \emph{yellow} are independent properties of animals.
In this case, the legal state space for the two variables contains all 4 possible configurations: 
\begin{equation}
S^o \defeq \cbr{(-1,-1), (-1,1), (1,-1), (1,1)}.
\end{equation}

\eat{
\begin{figure}[ht]
  \centering
  \includegraphics[width=2in]{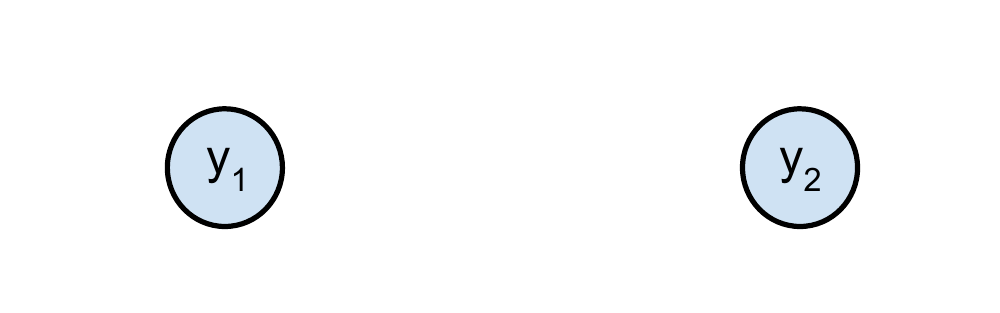}
  \caption{Non-relation in HEX graph: $y_1$, $y_2$ are mutually independent.}
\label{fig:hex-overlapping}
\end{figure}
}

\subsection{HEX graph as a graphical model}

To mathematically formulate the HEX model,
assume we have a  set of $n$ possible labels, represented as the bit vector $\yb = \cbr{y_1, \ldots, y_n}$,
where $y_i \in \{-1,+1\}$.
Also, assume we have an input feature vector 
$\xb = \cbr{x_1, \ldots, x_d}$, and some discriminative model which maps this to the score vector
$\zb=\cbr{z_1,\ldots,z_n}$, where $z_i$ is the ``local evidence'' for label $y_i$.
(The mapping from $\xb$ to $\zb$ is arbitrary; in this paper, we assume it is represented by a deep
neural network parameterized by $\wb$,
which we will denote by $\zb = DNN(\xb; \wb)$.) 
Given this, we can define the model as follows:
\begin{equation}
p(\yb|\xb) = \frac{1}{Z(\xb)}
  \prod_{i=1}^n \psi(y_i, z_i) 
\prod_{(i,j) \in G} \phi_a(y_i, y_j),
\label{eqn:crf}
\end{equation}
where $\psi(y_i, z_i) = 1/(1 + \exp(- 2 y_i z_i))$ is the logistic function,
and $\phi(y_i,y_j)$ is the (edge-specific) potential function, defined below:
(We use the notation $\phi_a$ to represent an ``absolute'' or
deterministic potential, to distinguish it from the soft or
probabilistic
potentials we use later, denoted by $\phi_p$.)
\begin{itemize}
\item Exclusion
\begin{align}
\phi_a^e(y_1,y_2) = \begin{cases}
1 & (y_1, y_2) \in S^e \\
0 & (y_1,y_2) = (1,1);
\end{cases}\label{eq:factor-ex}
\end{align}
\item Hierarchy
\begin{align}
\phi_a^h(y_1,y_2) = \begin{cases}
1 & (y_1, y_2) \in S^h \\
0 & (y_1,y_2) = (-1,1);
\end{cases}\label{eq:factor-sub}
\end{align}
\item No relation
\begin{align}
\phi_a^o(y_1,y_2) = 1 \;\; \forall (y_1,y_2).
\label{eq:factor-over}
\end{align}
\end{itemize}

\eat{
With these factor functions, we then define a binary-output function $\Phi_a^G(\yb)$ as a product of all factor functions in the graph to characterize a HEX graph $G$,
\begin{align}
\Phi_a^G(\yb) = \prod_{(i,j) \in G} \phi_a(y_i, y_j), \label{eq:hex-factor}
\end{align}
where $(i,j)$ denotes a directed or undirected edge in graph $G$, and the factor functions $\phi_a(y_i,y_j)$ comes from Equation \eqref{eq:factor-ex} or Equation \eqref{eq:factor-sub}. Because an overlapping relation corresponds to no-edge between the labels and its factor function is constant to 1, we can ignore that in Equation \eqref{eq:hex-factor}. When $\Phi_a^G(\yb) = 1$, the configuration $\yb$ is legal in $G$; otherwise when $\Phi_a^G(\yb) = 0$, the configuration $\yb$ is illegal. 

It is  worth noting that the HEX graph naturally generalizes two classical classification models:
binary multi-label logistic regression corresponds to a HEX graph with no edges; and 
the multiclass softmax model is a clique graph where all nodes are connected to each other by undirected edges. (This is up to an additional constant. See more discussions in Section \ref{sec:multinomial}.) 
}

\eat {
\subsection{Inference in HEX models}

HEX models have a few important properties which are crucial to enable efficient exact inference.
Let us define the product of all the potential functions:
\begin{align}
\Phi_a^G(\yb) = \prod_{(i,j) \in G} \phi_a(y_i, y_j)
\in \{0,1\}.
\label{eq:hex-factor}
\end{align}
We say that 
two HEX graphs $G$ and $G'$ are equivalent if and only if they have the same legal state space. 
In other words, 
 \begin{align*}
 \Phi_a^G(\yb) = \Phi_a^{G'}(\yb) \;\; \forall \yb. 
 \end{align*}
This allows us to sparsify and densify the HEX graph to its equivalence graph without changing the 
distribution $p(\yb|\xb)$.

\begin{figure}[ht]
\centering
\includegraphics[width=3in]{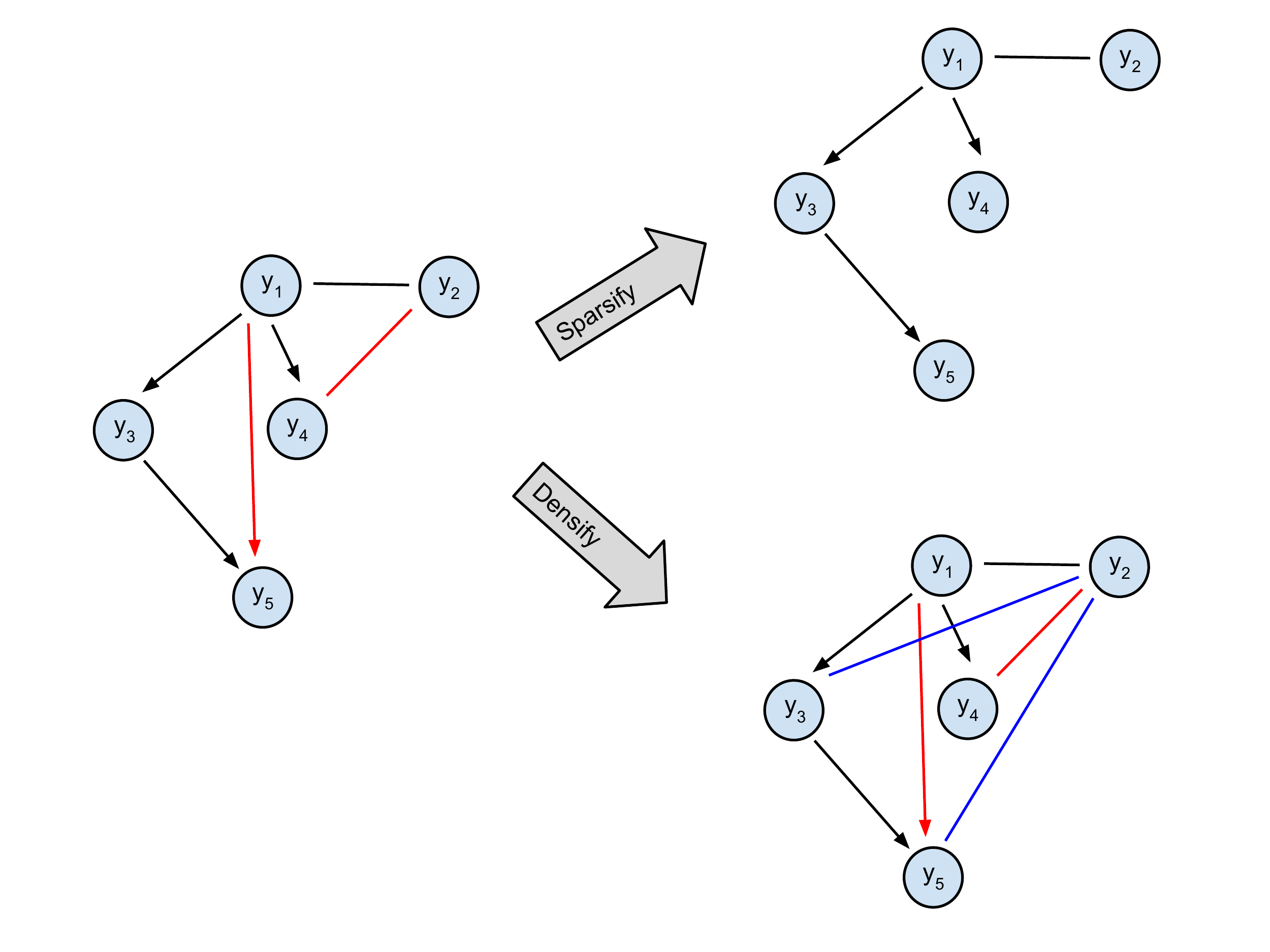}
\caption{An example of equivalent HEX graphs. }
\label{fig:equivalent-hex}
\end{figure}

Figure~\ref{fig:equivalent-hex} gives an example of three equivalent HEX graphs. In this example, exclusion of label $y_1$ and $y_2$ implies the exclusion of $y_4$ and $y_2$ because $y_1$ subsumes $y_2$,
hence we can drop the explicit undirected red $y_2 - y_4$ edge.
 Similarly, subsumption of $y_1$ to $y_3$ and $y_3$ to $y_5$ implies
 the subsumption of $y_1$ to $y_5$. Therefore, the left HEX graph can
 be sparsified to the top right graph without changing
 $\Phi(\yb)$. Similarly, we can also densify the left graph to obtain
 the bottom right graph by dropping the implicit directed blue edges.

Sparsification results in a sparse graph, which often reduces the treewidth of the resulting junction tree, resulting in faster inference. On the other hand, densification adds more constraints, thus reducing the size of the state space for each clique in the junction tree, again speeding up infernece. For example, if $n$ labels are connected to each other with an exclusive clique, then the the legal state space of the $n$ labels is only of size $n+1$ compared to the entire state space of size $O(2^n)$. For more details, see \cite{DenDinJia14}. 
}

\section{Probabilistic HEX models}
\label{sec:phex}

In this section, we introduce an extension of the HEX model to allow
for soft or probabilistic relationships between labels.
The basic idea is to relax the hard constraints, by replacing the
value 0  (corresponding to illegal
combinations) in the definitions of the potential functions with a value $0 \leq q \leq 1$,
representing how strongly we wish to enforce the constraints.
(This is somewhat  analogous to the approach  used in Markov logic
networks \cite{Richardson06}, which relax the hard constraints used in
first order logic.)
In principle, $q$ can be estimated from data along with the parameters of the unary potentials (discussed in Section~\ref{sec:learning}),
but in this paper, we either tie the $q$'s across all edges, or set them based on prior knowledge of the strength of the relations. 


\eat{
HEX graphs are a unified framework which effectively models absolute label relations by using legal and illegal states. In reality, only absolute relations are not sufficient in modeling many uncertain relations between objects and attributes. For example, in the U.S., a large number of taxi's are in yellow, but there are still some exceptions. It would be inappropriate to assign an absolute subsumption relation between "yellow" and "taxi", because it eliminates the possibility of taxi's in any other color. On the other hand, the two labels are not independent, therefore the absolute overlapping relation also does not apply. To enable accurate modeling of the rich semantic relations in reality, we have to rely on probabilistic relations. 
}

\subsection{Probabilistic HEX relations}

For clarity, we now explicitly specify the form of the two new factors
we introduce. We use the generic parameter $q$ to represent the
strength of this relation, although this could easily be made edge/
label dependent.

\paragraph{Probabilistic exclusion} 
The potential function of the two variables $y_1$, $y_2$ under probabilistic exclusion is defined as: 
\begin{align}
\phi_p^e(y_1,y_2; q) = \begin{cases}
1 & (y_1, y_2) \in S^e \\
q & (y_1, y_2) = (1,1),
\end{cases}\label{eq:factor-pex}
\end{align}
where $0 \le q \le 1$. When $q=1$, Equation \eqref{eq:factor-pex} reduces to
the non-relation in Equation \eqref{eq:factor-over}, where $y_i$
and $y_j$ are independent.
When $q=0$, Equation \eqref{eq:factor-pex} reduces to the hard exclusion relation Equation \eqref{eq:factor-ex}, where $(y_1,y_2)=(1,1)$ is strictly prohibited. 

\paragraph{Probabilistic hierarchy}
For hierarchy (subsumption), we define
\begin{align}
\phi_p^h(y_1,y_2; q) = \begin{cases}
1 & (y_1, y_2) \in S^h \\
q & (y_1,y_2) = (-1,1),
\end{cases}\label{eq:factor-psub}
\end{align}
where $0 \le q \le 1$. 
This reduces to
the unconstrained relation when $q=1$; and reduces to
the hard subsumption relation when $q=0$. 

\eat{
A pHEX graph uses undirected edges for probabilistic exclusions and directed edges for probabilistic subsumptions. When there is no-edge between two labels, they are considered to be independent (absolute overlapping). Similar to the HEX graph, we define a function $\Phi_p^G(\yb)$ for a pHEX graph $G$, 
$$\Phi_p^G(\yb) = \prod_{(i,j) \in G} \phi_p(y_i, y_j;q_{ij}), $$
with $\phi_p(y_i,y_j;q_{ij})$ being the probabilistic relation factor functions Equation \eqref{eq:factor-pex} and Equation \eqref{eq:factor-psub}. 

Unlike the HEX graph, the outcome of $\Phi_p^G(\yb)$ is not binary, but $\in [0,1]$. A larger $\Phi_p^G(\yb)$ indicates a more likely configuration while a smaller $\Phi_p^G(\yb)$ indicates a more susceptible configuration. 

\vspace{+0.1in}
A pHEX graph consists of both undirected and directed edges. In practice, it would be more straightforward to consider an undirected graphical model. Interestingly, it is possible to convert the pHEX graph to an equivalent Ising model. 
}

Probabilistic exclusions and subsumptions can be seen as a probabilistic mixture of absolute exclusions, subsumptions, and non-relations, where
\begin{align*}
\phi_p^e(y_1,y_2; q) &= q \phi_a^o(y_1,y_2) + (1-q) \phi_a^e(y_1,y_2),\\ 
\phi_p^h(y_1,y_2; q) &= q \phi_a^o(y_1,y_2) + (1-q) \phi_a^h(y_1,y_2). 
\end{align*}
Therefore, the combination of probabilistic label relations generalizes the absolute label relations in the HEX graph. 

\subsection{Converting pHEX models to Ising models}
\label{sec:ising}

The main disadvantage of this relaxation is that we lose the ability
to perform tractable exact inference. However, we now show that we can
formulate pHEX models as Ising models, which opens up the door to
using standard tractable approximate inference methods.


\begin{figure}[hpt]
  \centering




\includegraphics[width=3.5in]{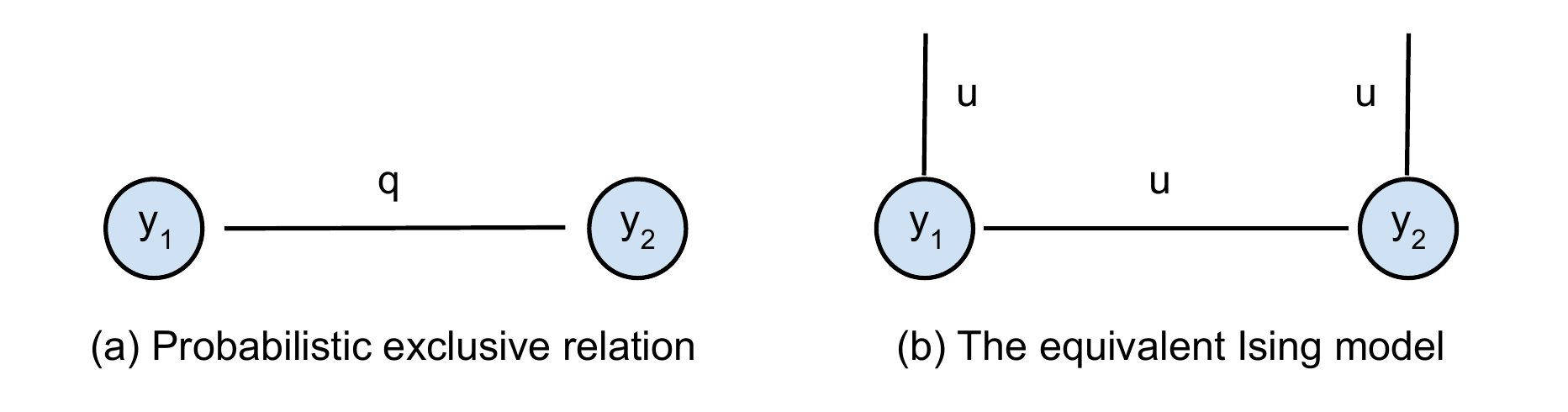}
  \caption{ (a) Probabilistic exclusive relations in a pHEX graph with $\phi(1,1) = q$; (b) the coefficients on the nodes and the edge of the equivalent Ising model, where $q = \exp(-4u)$. }
\label{fig:phex-exclusion}
\end{figure}

\begin{figure}[hpt]
  \centering
    



\includegraphics[width=3.5in]{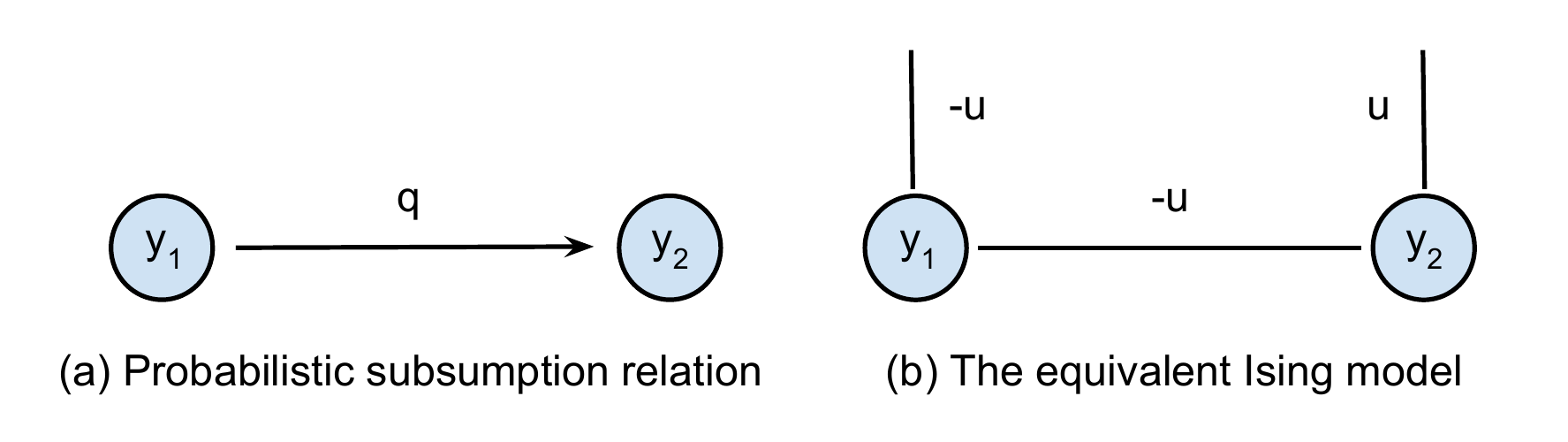}
  \caption{(a) Probabilistic subsumption relations in a pHEX graph  with $\phi(-1,1) = q$; (b) the coefficients of the equivalent Ising model, where $q = \exp(-4u)$. }
\label{fig:phex-subsumption}
\end{figure}

The Ising model was first proposed in statistical mechanics to study ferromagnetism \cite{Binder01}. Mathematically, it is essentially an undirected graphical model which defines the joint distribution of configurations of $n$ binary random variables $\yb$ in graph $G$ by a Boltzmann distribution, 
\begin{align}
p_\beta(\yb) = \frac{1}{Z_\beta} \exp(-\beta E(\yb)),
\end{align}
where  $Z_\beta$ is the normalization constant,
and $\beta$ is a temperature variable that will be omitted later by
fixing it to 1.
 $E(\yb)$ is the energy function of the configuration $\yb$, which takes into account local energy potentials $h_i y_i$ as well as pairwise energy potential $J_{ij} y_i y_j$,   
\begin{align}
E(\yb) = \sum_{(i,j) \in G} J_{ij} y_i y_j + \sum_{i=1}^n h_i y_i. 
\label{eq:ising}
\end{align}

To convert a pHEX graph to an Ising model, we first show how to convert the factor functions $\phi_p(y_1,y_2)$ for the pairwise probabilistic relations to the equivalent pairwise energy functions $E(y_1,y_2)$ of an Ising model.

Consider an Ising model of two variables in Figure \ref{fig:phex-exclusion}(b), where $u \ge 0$ are the weights on the local potentials and the pairwise potential. The resulting pairwise energy function of this Ising model is,
\begin{align}
E_p^e(y_1,y_2;u) &= u y_1 y_2 + u y_1 + u y_2\nonumber\\
&= \begin{cases}
-u & (y_1,y_2) \in S^e \\
3u & (y_1,y_2) = (1,1).
\end{cases}\label{eq:cost-pex}
\end{align}
Clearly, Equation \eqref{eq:cost-pex} looks very similar to
Equation \eqref{eq:factor-pex}.
In fact,  by letting $q=\exp(-4u)$ and
$\phi_p^e(y_1, y_2; q) \propto \exp(-E_p^e(y_1,y_2;u))$,
 we can show they are equivalent up
to a constant factor.
To see this, let $(y_1, y_2)$ be a legal label pair,
and $(y_1', y_2')$ be an illegal pair.
We have
\begin{align*}
\phi_p(y_1, y_2)/\phi_p(y_1',y_2') &= 1/q = e^{4u},\\
\exp(-E(y_1, y_2) + E(y_1',y_2')) &= e^{u + 3u} = e^{4u}.
\end{align*}
A larger $u$ means a stronger exclusion between the two labels. When
$u \to +\infty$, Equation \eqref{eq:cost-pex} reduces to the hard exclusive
relation; conversely when $u = 0$, Equation \eqref{eq:cost-pex} reduces to the
 non-relation.

Similarly, the equivalent Ising model of the probabilistic subsumption is shown in Figure \ref{fig:phex-subsumption}(b), where, 
\begin{align}
E_p^h(y_1,y_2;u) &= -u y_1 y_2 - u y_1 + u y_2\nonumber\\
&= \begin{cases}
-u & (y_1,y_2) \in S^h\\
3u & (y_1, y_2) = (-1,1).
\end{cases}
\label{eq:cost-psub}
\end{align}
We set
$\phi_p^h(y_1, y_2; q) \propto \exp(-E_p^h(y_1,y_2;u))$
and $q = \exp(-4u)$. 

The product of the pairwise factor functions $\phi_p(y_i,y_j;q_{ij})$
can now be written in terms of the sum of pairwise
energy functions $E(y_i,y_j;u_{ij})$:
\begin{align*}
&\prod_{(i,j) \in G} \phi_p(y_i,y_j,q_{ij})
\propto \exp \rbr{ -\sum_{(i,j) \in G} J_{ij} y_i y_j -\sum_{i=1}^n h_i y_i}, 
\end{align*}
where
\begin{align}
J_{ij} &= \begin{cases}
u_{ij}, & (i,j) \in ex.\\
-u_{ij}, & (i,j) \lor (j,i) \in sub.
\end{cases} \label{eq:ising-J}\\
h_i &= \sum_{\cbr{j|(i,j) \in ex.}} u_{ij} - \sum_{\cbr{k|(k,i) \in sub.}} u_{ki} + \sum_{\cbr{l|(i,l) \in sub.}} u_{il}. \label{eq:ising-h}
\end{align}
Here  $ex.$ denotes the set containing all exclusive relations
and $sub.$ the set containing all subsumption relations. Note that all
the pairs $(i,j) \in ex.$ satisfy $i < j$, and pairs $(i,j) \in
sub.$ means $i$ subsumes $j$.  

To incorporate local evidence into the model,
we can rewrite Equation \eqref{eqn:crf} as follows:
\begin{align*}
p(\yb | \zb) &\propto
\exp\rbr{\sum_{i=1}^n \log \psi(y_i,z_i) - \sum_{(i,j) \in G} E(y_i,y_j;u_{ij})}\\
 &= \exp(- \sum_{(i,j) \in G} J_{ij}y_i y_j -\sum_{i=1}^n (h_i - z_i) y_i),
\end{align*}
where $J_{ij}$ and $h_i$ are from Equation \eqref{eq:ising-J} and
Equation \eqref{eq:ising-h}. Note that we omitted a constant from the
$\log \psi$ term, because it will be canceled out
by the normalization constant $Z$.
By defining $h'_i = h_i - z_i$, we can ``absorb'' the local evidence
into the Ising model, and use standard inference methods.

\subsection{Inference in pHEX models}
\label{sec:inference}

At test time, we need to compute the marginal distribution per label, $p(y_i|\zb)$. In multi-label classification problems, a label $y_i$ is predicted to be true if $p(y_i | \zb) \ge 0.5$. In multi-class classification problems, the label
$$y^* = \argmax_{i=1}^n p(y_i | \zb)$$ is predicted to be the true label.
At training time, we need $p(y_i|\zb)$ as well as the term
$p(y_i|y_j = 1,\zb)$, where some of the true observed labels (\eg for node $j$) are set to their desired target states. 

\eat{
\begin{algorithm}[ht]
\SetKwInOut{Input}{Input}
\SetKwInOut{Output}{Output}

\caption{Testing Phase}
\Input{pHEX graph $G$, edge strength $q$, neural network parameters
  $\wb$, input features $\{\xb^{(b)}\}$
}
\Output{$p(y_i|\xb^{(b)})$ for all labels $i$ and examples $b$}
Compute $J_{ij}$, $h_i$ using Equation \eqref{eq:ising-J}, \eqref{eq:ising-h}\;
\For {$b = 1, 2, \ldots $} {
$\zb^{(b)} = DNN(\xb^{(b)}; \wb)$ \;
Update local potentials of pHEX graph with $h_i' = h_i - z_i^{(b)}$ for all $i$\;
Run LBP to obtain $p(y_i|\zb^{(b)})$ for all $i$\;
Output $p(y_i|\zb^{(b)})$ for prediction.
}
\label{alg:phex-test}
\end{algorithm}
}

Exact inference in pHEX models is usually intractable, when the graphs are loopy, and the legal states 
are not sparse. Since $p(\yb|\zb)$ is an Ising model, 
we can apply any
off-the-shelf inference method,
including mean-field inference (MF), loopy belief propagation (LBP), and Markov Chain Monte Carlo (MCMC) methods \cite{Bishop2006}. In practice, we find that the standard LBP algorithm works consistently well, so we use it as our main inference algorithm in our experiments. We give the details below.


We define the belief on each label $y_i$ to be $b_i(-1)$ and $b_i(1)$, and the message from $y_i$ to its neighbour $y_j$ to be $m_{i \to j}(-1)$ and $m_{i \to j}(1)$. Then the algorithm iterates through all beliefs and messages with updates, 
\begin{align*}
b_i(1) &\propto \exp(-h'_i) \prod_{j \in N(i)} m_{j \to i}(1), \\
b_i(-1) &\propto \exp(h'_i) \prod_{j \in N(i)} m_{j \to i}(-1), 
\end{align*}
where $N(i)$ denotes the neighbours of $i$, and  
\begin{align*}
m_{j \to i}(1) &\propto \exp(-J_{ij}) \frac{b_i(1)}{m_{i \to j}(1)} + \exp(J_{ij}) \frac{b_i(-1)}{m_{i \to j}(-1)},\\
m_{j \to i}(-1) &\propto \exp(-J_{ij}) \frac{b_i(-1)}{m_{i \to j}(-1)} + \exp(J_{ij}) \frac{b_i(1)}{m_{i \to j}(1)}.
\end{align*}

To maintain numerical stability, we normalize $b_i$ and $m_{j \to i}$
throughout inference,
and we perform updates in the log domain.
After all beliefs have converged or a maximum number of iterations has
been reached, we estimate the marginal probabilities  by $p(y_i = 1|\zb) = b_i(1)$.

The inference of $p(y_i|y_j = 1,\zb)$ is almost the same as above
except we set $b_j(1)=1$ and $b_j(-1)=0$ to represent the fact that node $j$ is clamped to state 1.
(We can easily extend this procedure if we  have multiple clamped nodes.)


\section{Mutually exclusive and collectively exhaustive relations}
\label{sec:multinomial}
\eat{
Sometimes we have some hard constraints as well as soft constraints.
In particular, the multiclass softmax relation has $k$ mutually exclusive binary variables which are also collectively exhaustive (MECE): 
\begin{align*}
\sum_{i=1}^k y_i = 2 - n, \;\; \forall y_i \in \cbr{-1, 1}. 
\end{align*}
HEX graph uses a clique with exclusive relations among all $k$ labels to approximate the MECE relation, but it introduces an unnecessary constant representing ``none of above''. }

In addition to allowing soft relations, our pHEX framework offers another advantage over HEX graphs: it is easy to enforce a new type of constraint, namely Mutually Exclusive and Collectively Exhaustive (MECE) relations, used in the multi-class softmax model. In HEX graphs, there is no way to express the notion of ``collectively exhaustive'', \ie, one of the mutually exclusive classes must be true. HEX graph thus has to maintain an additional ``none of the above'' state.

In the pHEX graph, we handle the MECE relation of $k$ nodes using a single multinomial variable with $k$ possible states. 
Although an undirected graphical model with multinomial nodes is strictly speaking not an Ising model, a slight variant on the standard LBP algorithm can still be applied for efficient approximate inference. 

For simplicity, we only illustrate the inference algorithm for pHEX graphs with one multinomial label node, since this will be used in later experiments. Further generalization to pHEX graphs with multiple multinomial nodes is straightforward and follows similar procedures. 

Let us denote the multinomial node by $c = \cbr{c_1, \ldots, c_k}$. The node and message updates for the standard binary nodes are the same  as
before. The belief of the multinomial node $c$ is updated as, 
\begin{align*}
b_{c}(i) \propto \exp(-h'_i) \prod_{j \in N(c)} m_{j \to c}(i) 
\end{align*}
for state $i \in \cbr{1, \ldots, k}$ in which $y_{c_i}=1$. Here $N(c) = \cup_{i=1}^k N(c_i)$ is the neighbour set of the multinomial node. The message from a standard node $j$ to the multinomial node $c$ is, 
\begin{align*}
m_{j \to c}(i) \propto& \exp(\sum_{s=1}^k J_{jc_s}- 2 J_{jc_i})\frac{b_j(1)}{m_{c \to j}(1)} \\
&\quad + \exp(-\sum_{s=1}^k J_{jc_s} + 2 J_{jc_i})\frac{b_j(-1)}{m_{c \to j}(-1)}
\end{align*}
for state $i$. The message from the multinomial node to a standard node $j$ is, 
\begin{align*}
m_{c \to j}(1) \propto& \sum_{i=1}^k \exp(\sum_{s=1}^k J_{jc_s}- 2 J_{jc_i})\frac{b_c(i)}{m_{j \to c}(i)},\\
m_{c \to j}(-1) \propto& \sum_{i=1}^k \exp(-\sum_{s=1}^k J_{jc_s}+ 2 J_{jc_i})\frac{b_c(i)}{m_{j \to c}(i)}.
\end{align*}
As in the standard LBP algorithm, we normalize $b_c$, $m_{j \to c}$ and $m_{c \to j}$ and update them in the log domain.
After the algorithm converges, the marginal probability of a node $c_k$ in clique $c$ is $p(y_{c_k}=1|\zb) = b_c(k)$.

\section{Learning}
\label{sec:learning}

An important property of the (p)HEX model is that not all the target labels need to be specified during training. For example, consider a data set of images. It is more common for a user to use basic level category names, such as ``dog'', than very specific names such as ``husky'' or ``beagle''. Furthermore, a user may not label everything in an image.
So the absence of a label is not evidence of its absence.

To model this, we allow some of the labels to be unobserved or hidden during training.
For example, if we clamp the ``husky'' label to true, and leave all other label nodes unclamped, the hard constraints will force the ``dog'' label to turn on, indicating that this instance is an example of both the husky class and the dog class.
However, if we clamp the ``dog'' label to true, we will not turn on ``husky'' or ``beagle'', since the relation is asymmetric.
We can also clamp labels to the off state, if we know that the corresponding class is definitely absent. For example, turning on ``dog'' will turn off ``cat'' if they are mutually exclusive. (In the pHEX case, the ``illegal'' states are down weighted, rather than given zero probability.)

Let the input scores for the $b$'th training instance be $\zb^b$,
and let the subset of target labels be $\tb^b=(t_1^b, \ldots, t_m^b)$,
where we have assumed that $m$ labels are observed in every instance for notational simplicity.
A natural loss function is the negative log likelihood of the observed labels
given the inputs:
\begin{align*}
L &= -\sum_{b=1}^N \sum_{j=1}^m \log p(y^b_{t_j}=1 | \zb^b). 
\end{align*}

\eat{
\begin{figure}[ht]
\centering
\includegraphics[width=2.5in]{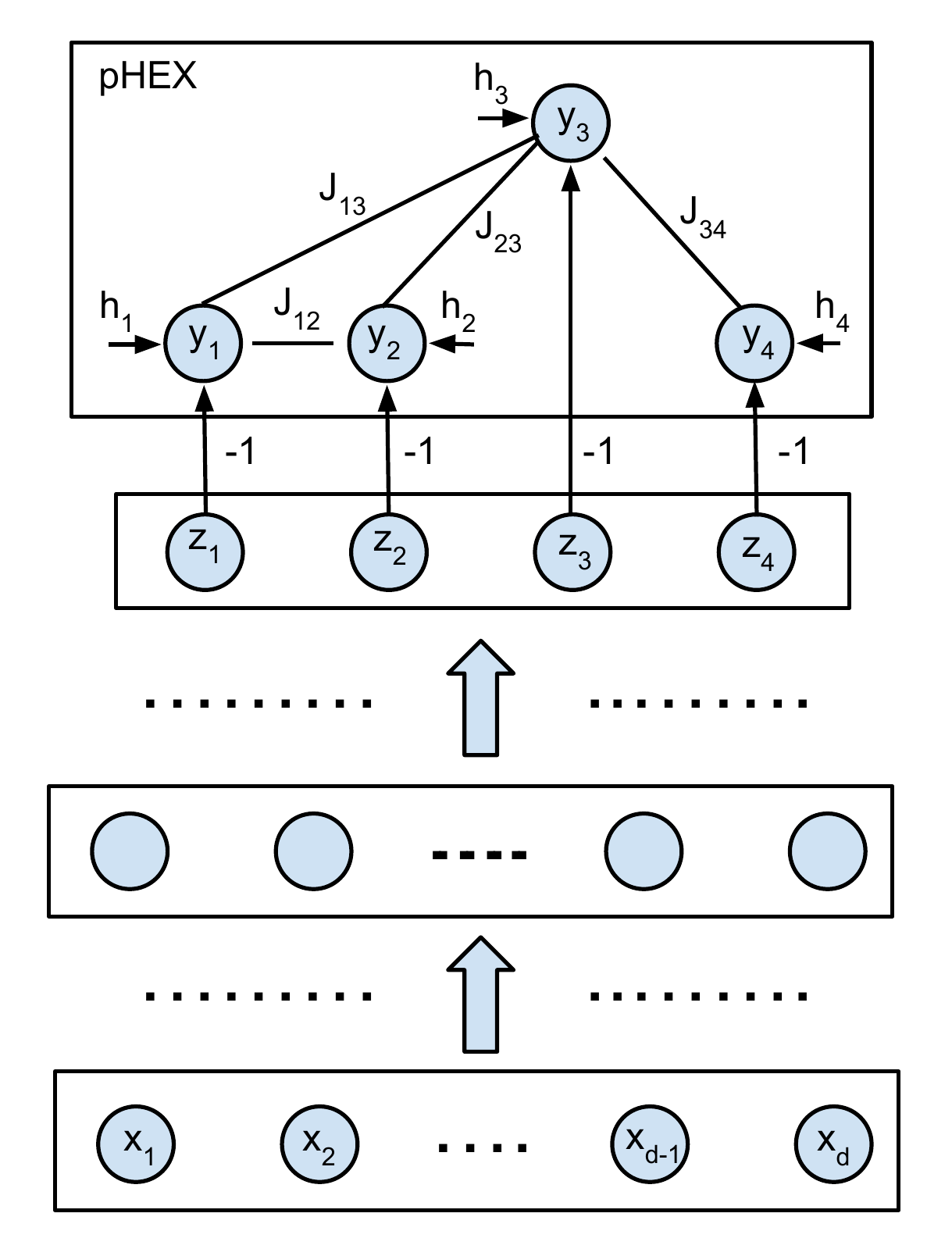}
\caption{The pHEX graph as a top layer of a deep feed-forward neural network. }
\label{fig:deepnets-phex}
\end{figure}

\subsection{Backpropagating the error to the local classifers}
}

To fit the local classifiers (unary potentials),
we first need to derive the gradient of the loss wrt the input scores $z_i$.
The derivative of $\log p(y_{t_j}=1 | \zb)$ over some $z_i$ is, 
\begin{align*}
&\frac{\partial \log p(y_{t_j}=1 | \zb) }{\partial z_i} 
= \EE_{p(y_i|y_{t_j} = 1,\zb)}[y_i] - \EE_{p(y_i|\zb)}[y_i].  
\end{align*}
Therefore, we need to compute the
conditional distributions $p(y_i|y_{t_j}=1, \zb)$ 
and  marginal distributions $p(y_i|\zb)$ for all $i$.
These correspond to the well-known ``clamped'' and ``unclamped''
phases of MRF / CRF learning.
We can then backpropagate the gradient into the parameters of the
local classifiers themselves.

We can use a similar gradient-based training scheme to estimate the
CRF edge parameters. However, in this paper, we simply
combine prior edge weights from data with a one-dimensional grid search of rescaling factor.

\eat{
\subsection{Details of the learning algorithm}

At the beginning of training, the pHEX graph of the candidate labels is converted to its equivalent Ising model with coefficients $J_{ij}$ and $h_i$ using Equation \eqref{eq:ising-J} and Equation \eqref{eq:ising-h} which are stored inside the pHEX layer and
used throughout the training.  During training, each training example's inputs $\xb = \cbr{x_1, \ldots, x_d}$ forward propagates through the feed-forward network and outputs $\zb$ as the inputs of the pHEX graph. The size of $\zb$ is the same as the size of $\yb$. Each $y_i$ takes $-z_i$ as the input and combines it with the pre-stored $h_i$ according to Equation \eqref{eq:classification-exp} to obtain the local potential of the resulting Ising model for inference.

The output of the pHEX layer is different in training and testing phases. During the training phase, we use stochastic gradient descent to minimize the loss function Equation \eqref{eq:phex-loss}. To this end, we first perform an LBP inference to estimate $p(y_i|y_{t_j} = 1,\zb)$ and $p(y_i|\zb)$, and compute the derivative $\partial L / \partial \zb$ based on Equation \eqref{eq:phex-grad}. Then we back-propagate the derivative through the $z_i$'s to the whole network. Therefore, the output of the pHEX layer during training is $\partial L / \partial z_i$ to each $z_i$. The training algorithm of the entire system is summarized in Algorithm \ref{alg:phex-train}. 

\begin{algorithm}[ht]
\SetKwInOut{Input}{Input}
\SetKwInOut{Output}{Output}

\caption{Training Phase}
\Input{pHEX graph $G$, edge strength $q$, labeled data $\{\xb^{(b)}, \tb^{(b)}\}$}
\Output{Neural network parameters $\wb$}
Initialize $\wb = \wb^{(0)}$\; 
Compute $J_{ij}$, $h_i$ using Equation \eqref{eq:ising-J}, \eqref{eq:ising-h}\;
\For {$b = 1, 2, \ldots $} {
$\zb^{(b)} = DNN(\xb^{(b)}; \wb)$ \;
Update local potentials of pHEX graph with $h_i' = h_i - z_i^{(b)}$ for all $i$\;
Run LBP to obtain $p(y_i|\zb^{(b)})$ for all $i$\;
Run LBP to obtain $p(y_i|y_{t_k}, \zb^{(b)})$ for all $k$ and $i$\;
Evaluate $\partial L/\partial \zb$ using Equation \eqref{eq:phex-grad} and output to $\zb^{(b)}$\;
Back propagate the gradient from $\zb^{(b)}$ to $\xb^{(b)}$ and update $\wb$\;
}
\label{alg:phex-train}
\end{algorithm}
}

\section{Experiments}
\label{sec:experiment}

In \cite{DenDinJia14}, the HEX graphs shows significant improvement over standard softmax and (multi-label) logistic regression models, so in this paper, we will just compare pHEX to HEX. We conduct three experiments.

\begin{figure}[htp]
\centering
\includegraphics[width=1.3in]{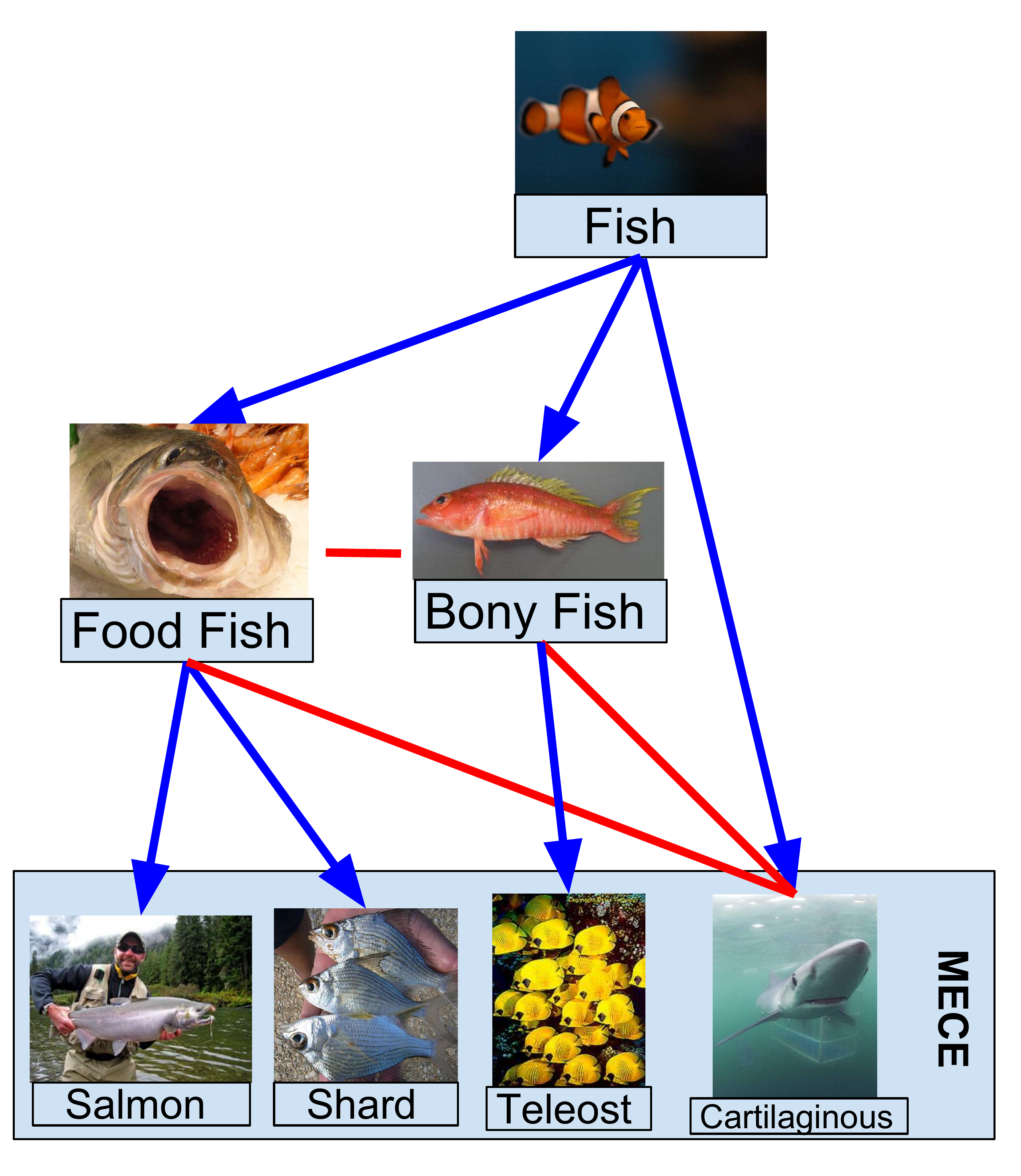}
\hspace{+0.3in}
\includegraphics[width=1.3in]{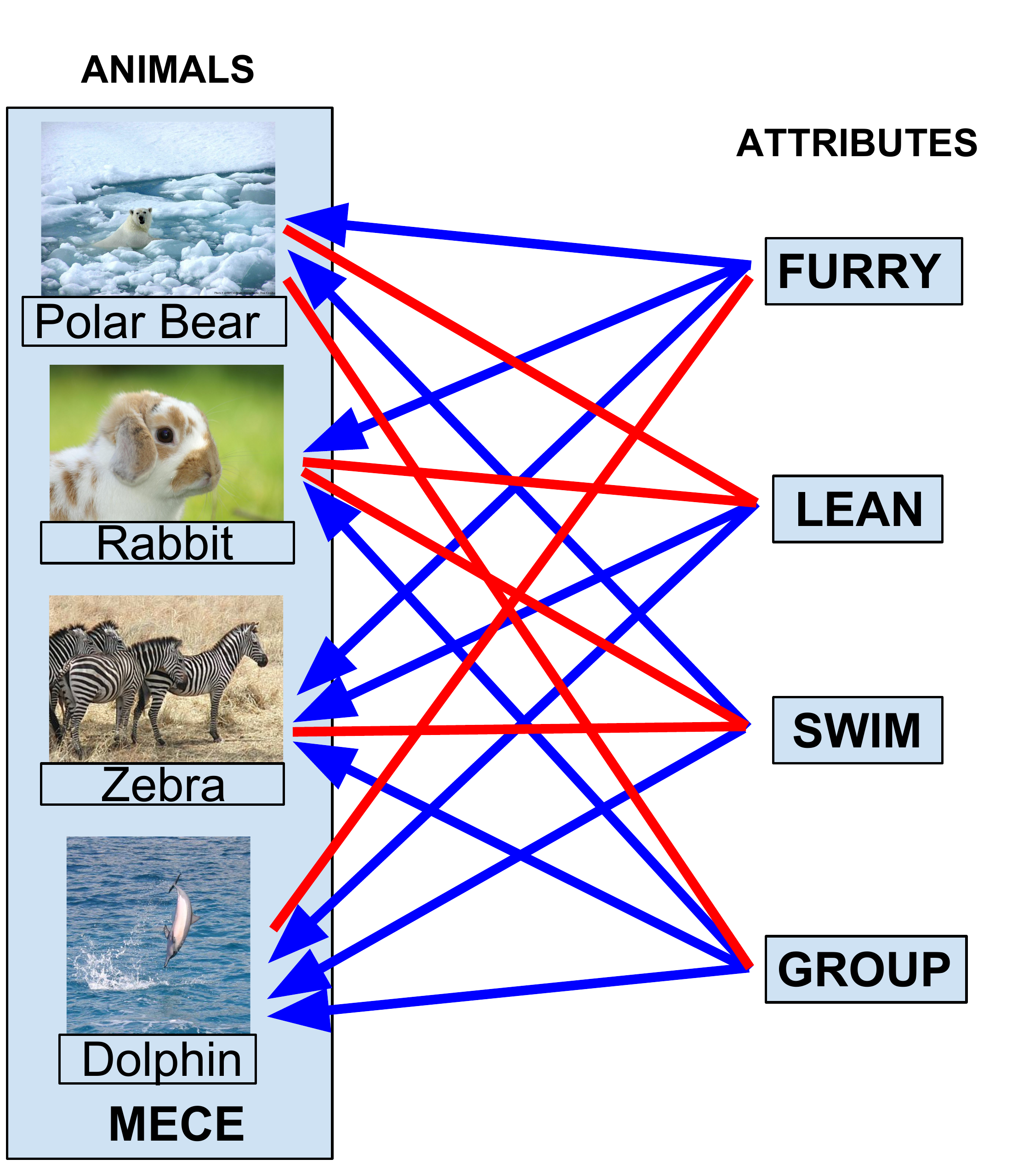}
\caption{Left: An illustration of the (p)HEX graph based on the WordNet hierarchy in the ImageNet experiments. Right: An illustration of the (p)HEX graph in the Animal with Attributes experiments.  The blue directed edges denote the subsumption relations; and the red undirected edges denote the exclusive relations. An MECE relation (multinomial node) is placed in the final pHEX graph. }
\label{fig:phex-cartoon}
\end{figure}

The first experiment is the standard ImageNet image classification problem
\cite{deng2012}.
We add hierarchical relations between the labels based on
the publicly available WordNet hierarchy. Since WordNet does not have exclusive relations, we assume that any two labels are exclusive if they are not in subsumption relation. Figure \ref{fig:phex-cartoon} Left is an example of the subgraph of "fish". As in \cite{DenDinJia14}, we assume that
the training labels are drawn from different levels of the hierarchy.
In this paper, we show
that pHEX with (constant) soft relations improves on HEX, especially when leaf labels are rarely present in the training set.

The second experiment is a zero-shot learning task,
in which we must predict unseen classes at test time, leveraging known relations between the class labels and attributes of the class. We use the Animals with Attributes dataset \cite{lampert2009}.
Following \cite{DenDinJia14},
we first assume that all object classes are mutually exclusive. We then add subsumption relations from a predicate (or attribute) to an object if the binary predicate of the object is 1, and add exclusive relations between predicate and objects if the binary predicate of the object is 0. See the illustration in Figure \ref{fig:phex-cartoon} Right.
In this paper, we relax the hard constraints and show that pHEX can work significantly better than HEX.
Finally, the third experiment is another zero-shot learning task, this time on the 
PASCAL VOC/ Yahoo images with attributes dataset \cite{Farhadi09}.
Again, we show that pHEX can significantly outperform HEX.
%



\subsection{Experimental setup}
\label{sec:exp-setup}

\begin{table}[htp]
  \caption{The Ising coefficients $u$ as well as the corresponding strengths of the label relations $q$ used in pHEX graphs in the experiments, where $q = \exp(-4u)$.}
  \begin{center}
    \begin{tabular}{||c||c|c|c|c|c|c|c||}
      \hline
      u & 0 & 0.1 & 0.3 & 0.5 & 0.7 & 1.0 & 1.5 \\ \hline
      q & 1 & 0.67 &0.30 &0.14 &0.06 &0.02 &0.002\\
      \hline
    \end{tabular}
  \end{center}
\label{table:ising_coefficient}
\end{table}

In our experiments, we used two types of pHEX graphs.
For the ImageNet experiments, we use the same constant edge strength
for all edges; we
vary this edge paramter $u$ across the ranges shown in
Table~\ref{table:ising_coefficient},
and plot results for each value.
For the zero-shot experiments, we consider constant edge weights,
but we also consider variable edge weights, which
we derive by scaling the prior edge weight (derived from the data)
by a global scale factor $u$, which we again
vary across a range.

\eat{
Since pHEX graphs are generalizations of the HEX graphs, a natural way to build a pHEX graph is based on an existing HEX graph. In our experiment, we build our pHEX graph by incorporating a constant Ising coefficinet $u$ on all edges of the HEX graph with maximal sparsification. Especially, to study the pHEX graphs with different strengths of label relations, we varied different Ising coefficients as shown in Table \ref{table:ising_coefficient}.
}

Note that, since all three tasks are evaluated on test labels in a multi-class setting, we add a MECE relation into the pHEX graphs
In particular, for the ImageNet dataset, 
we add a multinomial node on the 1000 leaf labels; in the Animal with Attributes dataset, we add a multinomial node on the 50 animal classes; and in the VOC/Yahoo dataset, we add a multinomial node on the 32 object classes. After adding MECE relations, we remove the replicated soft exclusive relations from the pHEX graph.

\subsection{ImageNet classification experiments}
\label{sec:imagenet}

\begin{figure*}[htp]
\centering
\includegraphics[width=1.5in]{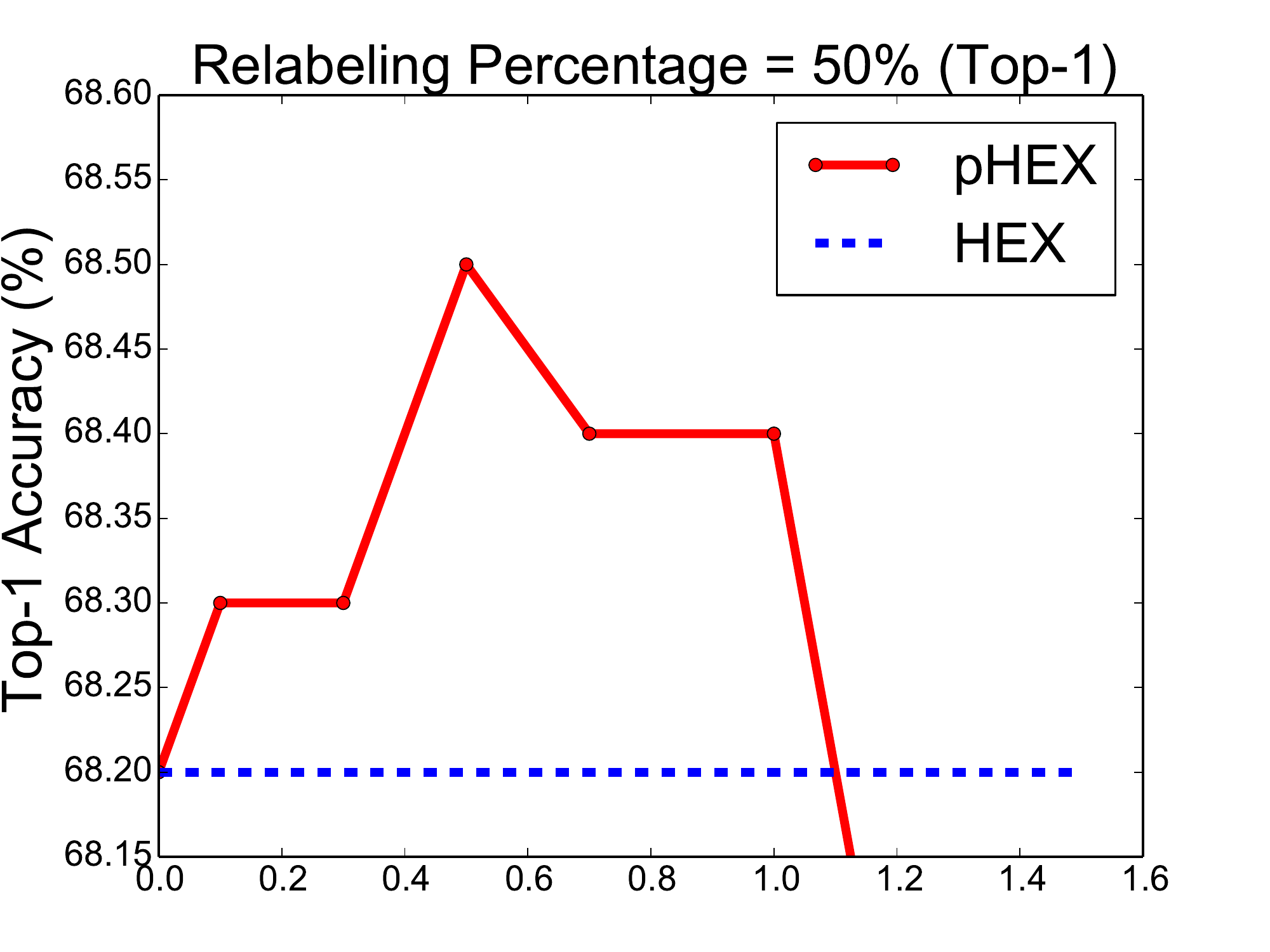}
\includegraphics[width=1.5in]{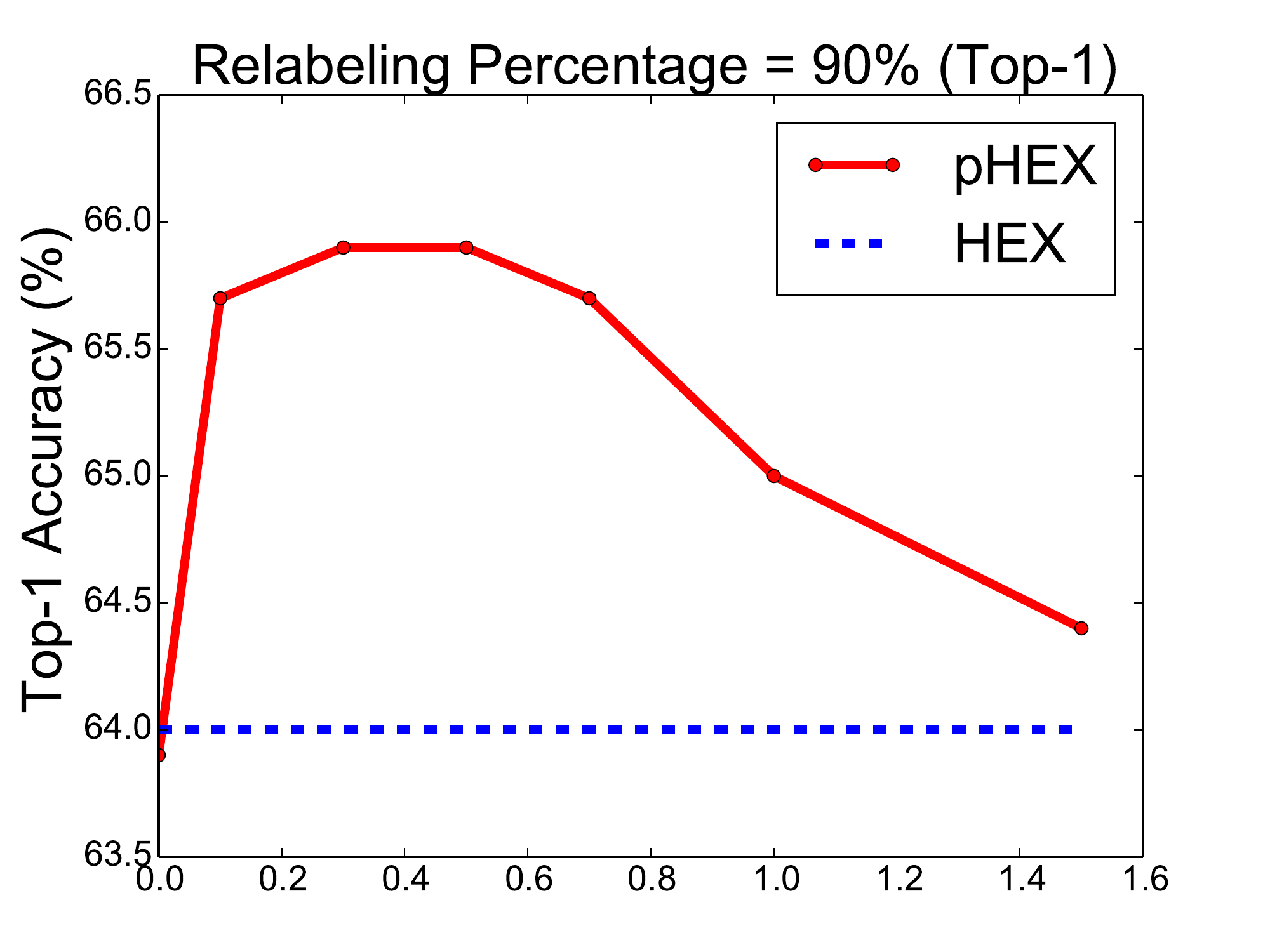}
\includegraphics[width=1.5in]{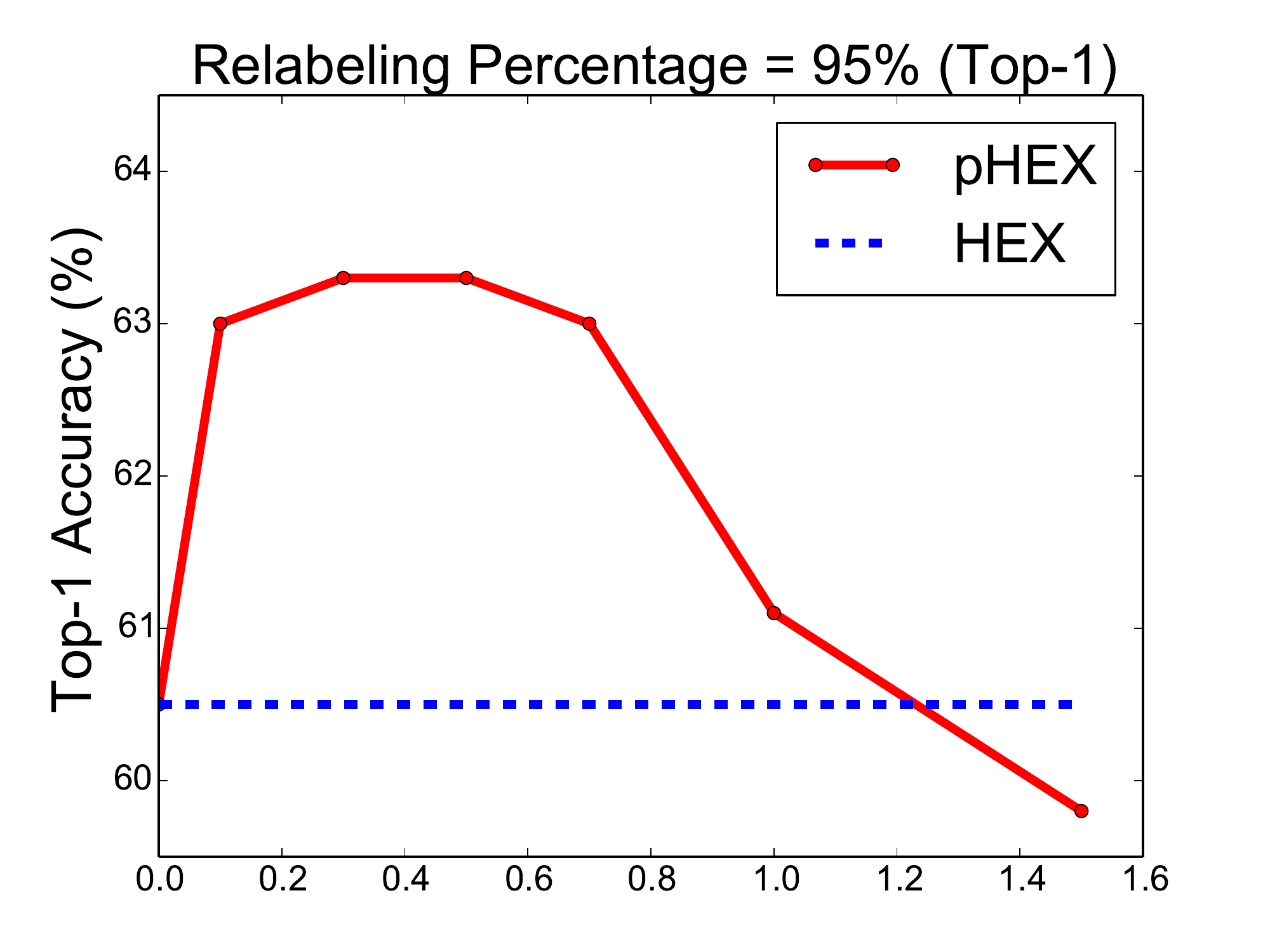}
\includegraphics[width=1.5in]{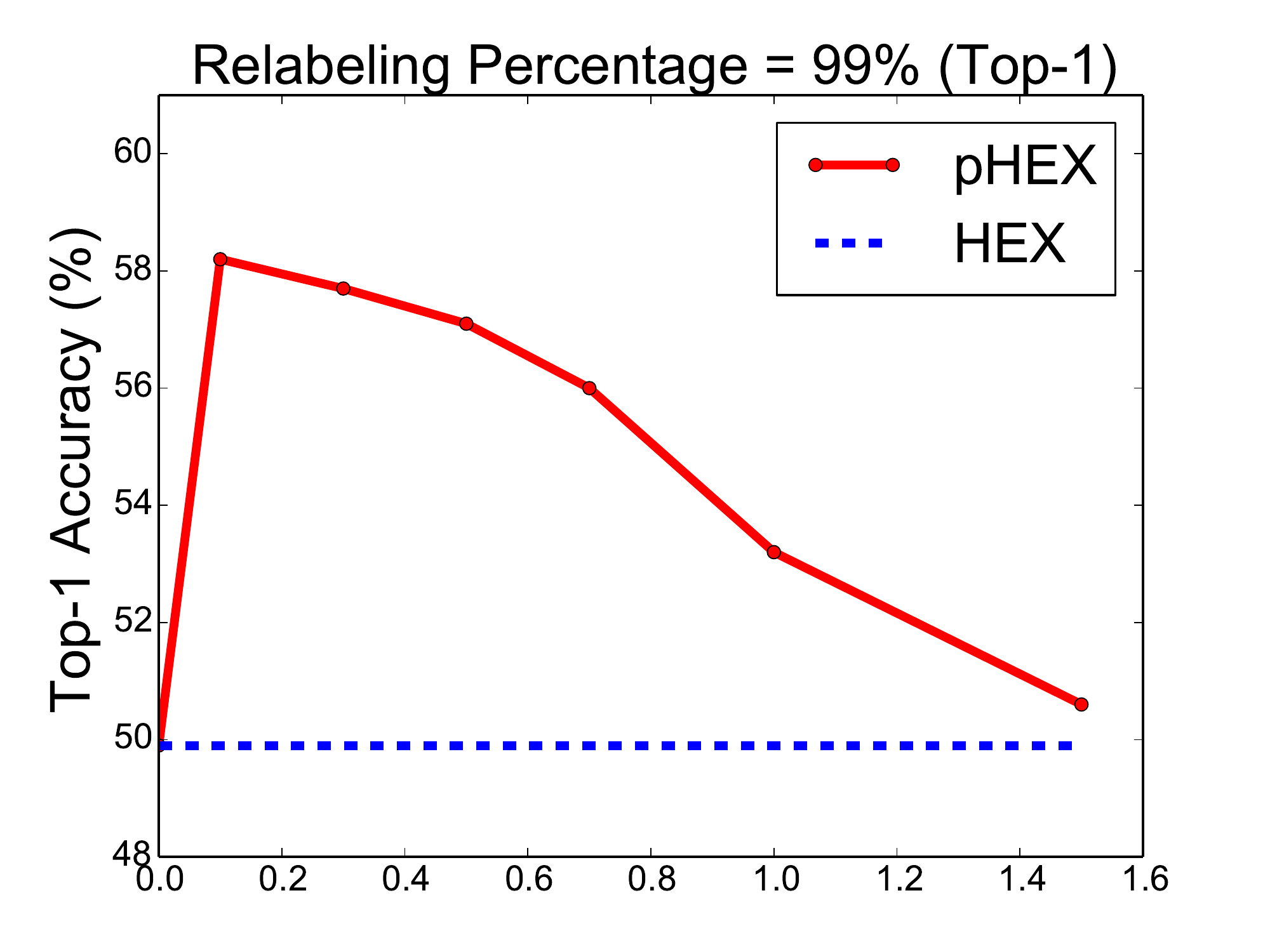}

\includegraphics[width=1.5in]{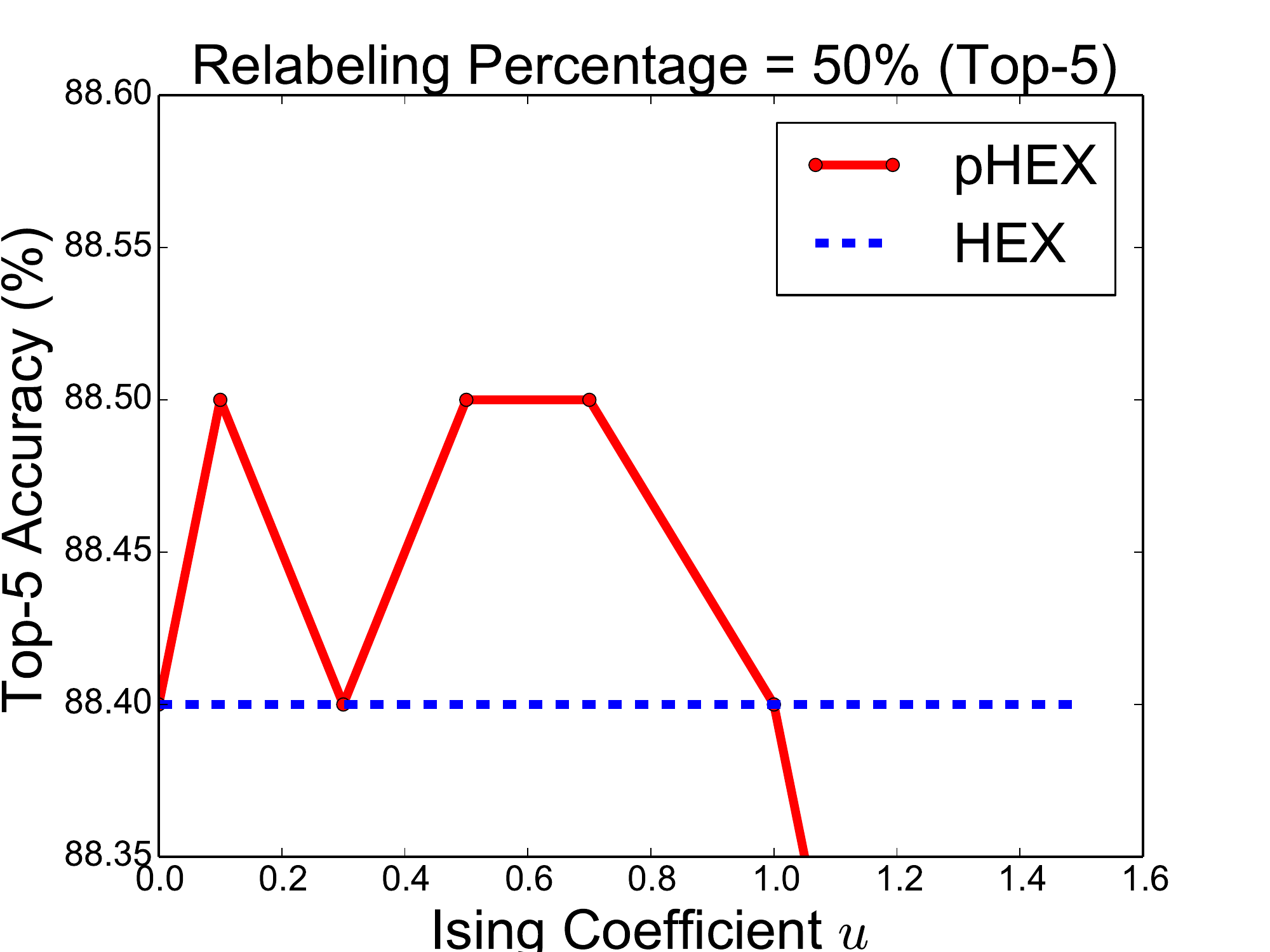}
\includegraphics[width=1.5in]{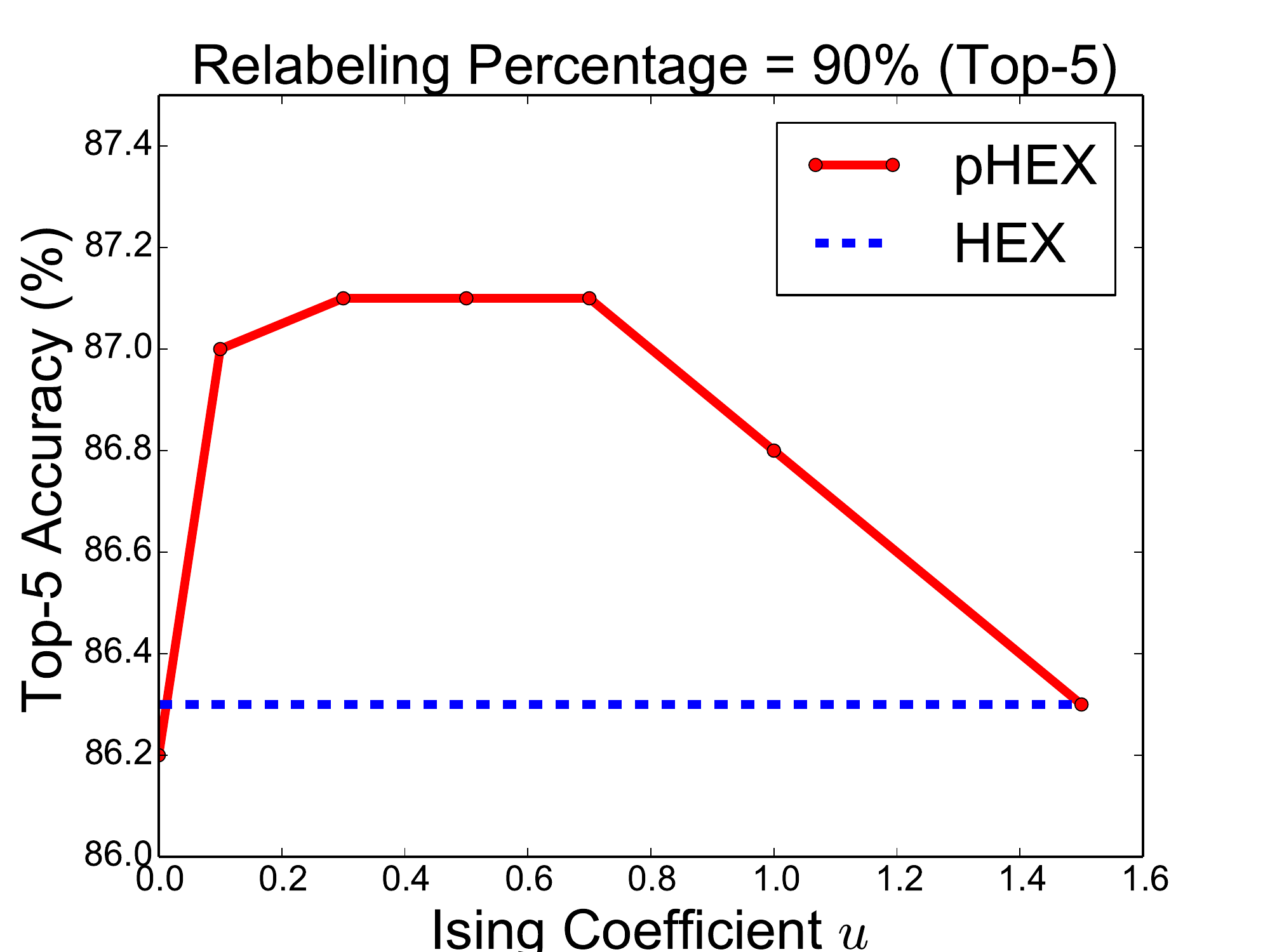}
\includegraphics[width=1.5in]{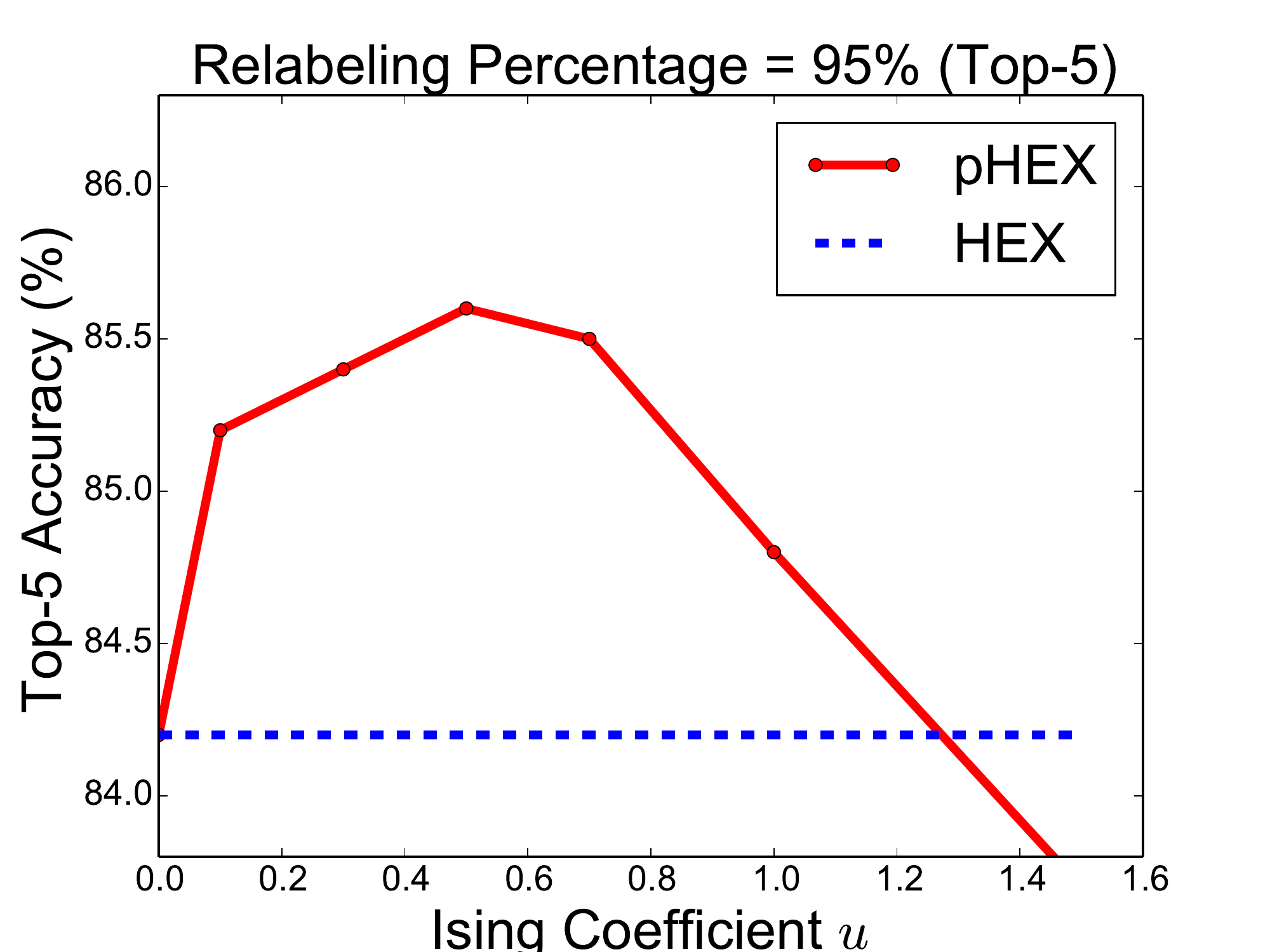}
\includegraphics[width=1.5in]{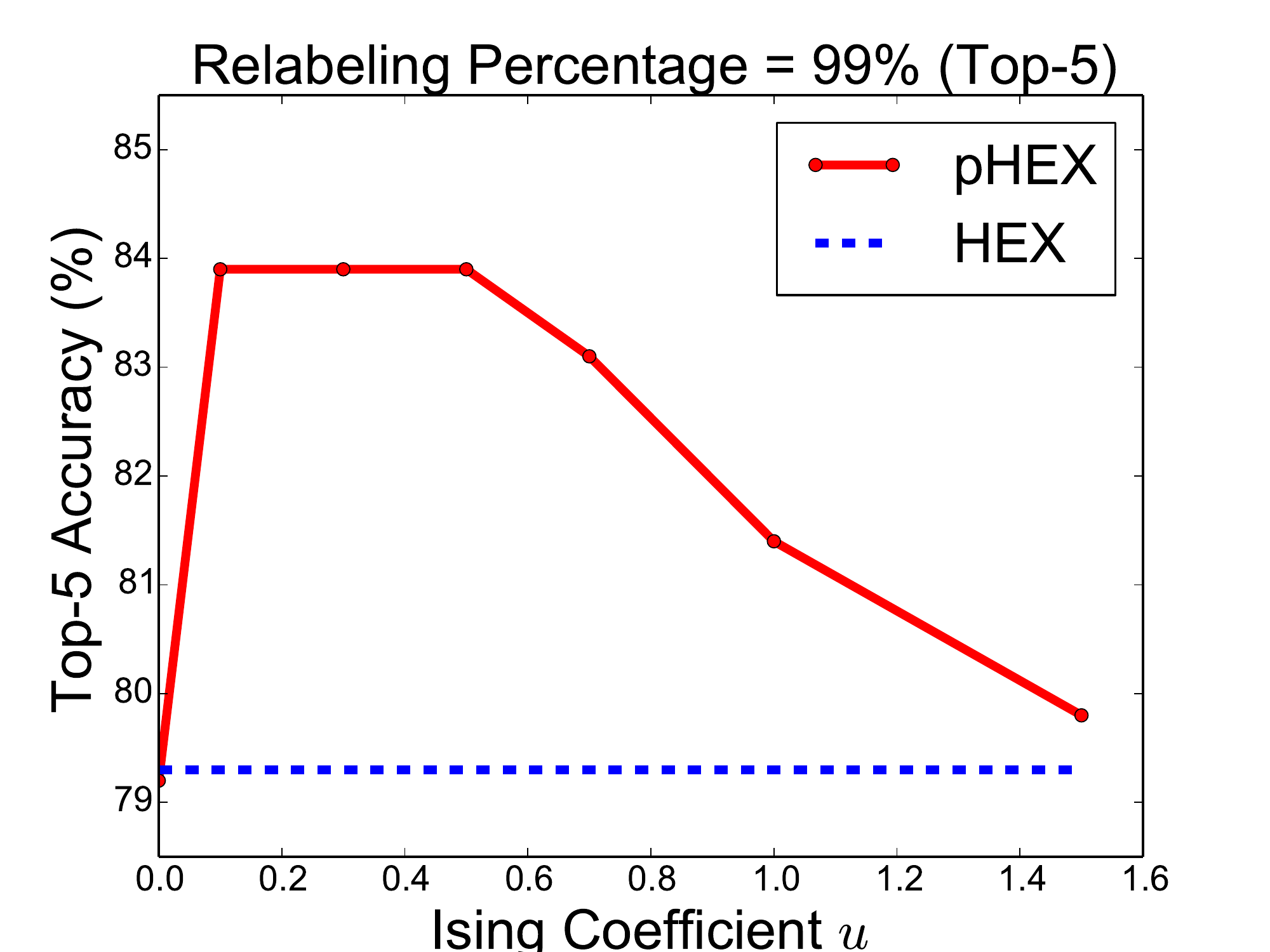}
\caption{Top-1 (top) and Top-5 accuracies (bottom) vs relation strength $u$ for the ImageNet classification experiment. The results of the pHEX graphs are in the red solid curves, and the results of the HEX graphs are in the blue dashed horizontal lines. 
From left to right: relabeling 50\%, 90\%, 95\%, 99\%.
}
\label{fig:imagenet}
\end{figure*}

In this section, we use the ILSVRC2012 dataset \cite{deng2012}, which consists of 1.2M training images from 1000 object classes. These 1000 classes are mutually exclusive leaf nodes of a semantic hierarchy based on WordNet that has 860
internal nodes. As in \cite{DenDinJia14}, we evaluate the recognition performance in the multiclass classification at the leaf level, but allow the training examples to be labeled at different semantic levels. Since ILSVRC2012 has no training examples at internal nodes, we create training examples for internal nodes by relabelling $\cbr{50\%, 90\%, 95\%, 99\%}$ of the leaf examples to their immediate parents based on the WordNet Hierarchy. Since the ground truth for test set is not released for ILSVRC2012, we use 10\% of the released validation set as our validation set and the other 90\% as our test set.

The underlying feed-forward network that we use is based on a deep convolutional neural network GoogLeNet \cite{Szegedy2014}. Since GoogLeNet  is such a large model, we adopt the following staged training procedure.
First we pre-train a CNN with a HEX graph as the top layer until convergence. Then we fine tune the entire model with pHEX graph layers of different coefficients $u$ on top. This can be thought of as a form of curriculum learning \cite{BenLouCol2009} by training with a simpler model (HEX graph) with exact inference first.

\eat{
\begin{table*}[htp]
  \caption{Top 1 (top 5 in brackets) classification accuracy on 1000 classes of
ILSVRC2012 with relabeling of leaf node data to internal nodes during training.}
  \begin{center}
    \begin{tabular}{||c||c|c|c|c||}
      \hline
      Relabeling & 50\% & 90\% & 95\% & 99\% \\
      \hline
	$u=0.1$ & 68.3 {\bf (88.5)} & 65.7 (87.0) & 63.0 (85.2) & {\bf 58.2 (83.9)} \\
	$u=0.3$ & 68.3 (88.4) & {\bf 65.9 (87.1)} & {\bf 63.3} (85.4) & 57.7 {\bf (83.9)} \\
	$u=0.5$ & {\bf 68.5 (88.5)} & {\bf 65.9 (87.1)} & {\bf 63.3 (85.6)} & 57.1 {\bf (83.9)} \\
	$u=0.7$ & 68.4 {\bf (88.5)} & 65.7 {\bf (87.1)} & 63.0 (85.5) & 56.0 (83.1) \\
	$u=1.0$ & 68.4 (88.4) & 65.0 (86.8) & 61.1 (84.8) & 53.2 (81.4) \\
	$u=1.5$ & 67.4 (87.9) & 64.4 (86.3) & 59.8 (83.7) & 50.6 (79.8) \\
      \hline
      HEX & 68.2 (88.4) & 64.0 (86.3) & 60.5 (84.2) & 49.9 (79.3) \\
	\hline
    \end{tabular}
  \end{center}
  \label{table:imagenet}
\end{table*}
}


\eat{
\begin{figure}[htp]
\centering
\includegraphics[width=1.5in]{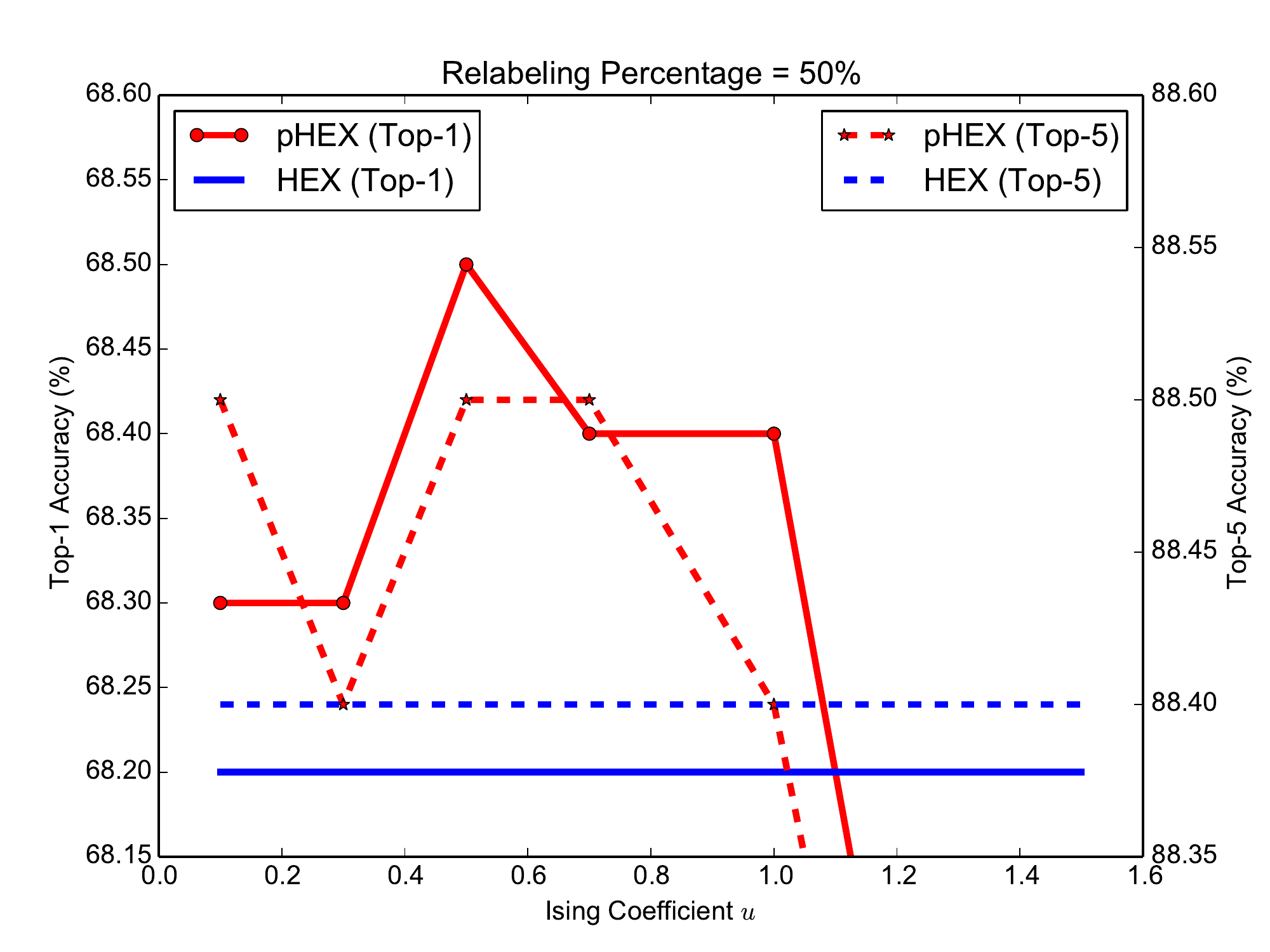}
\includegraphics[width=1.5in]{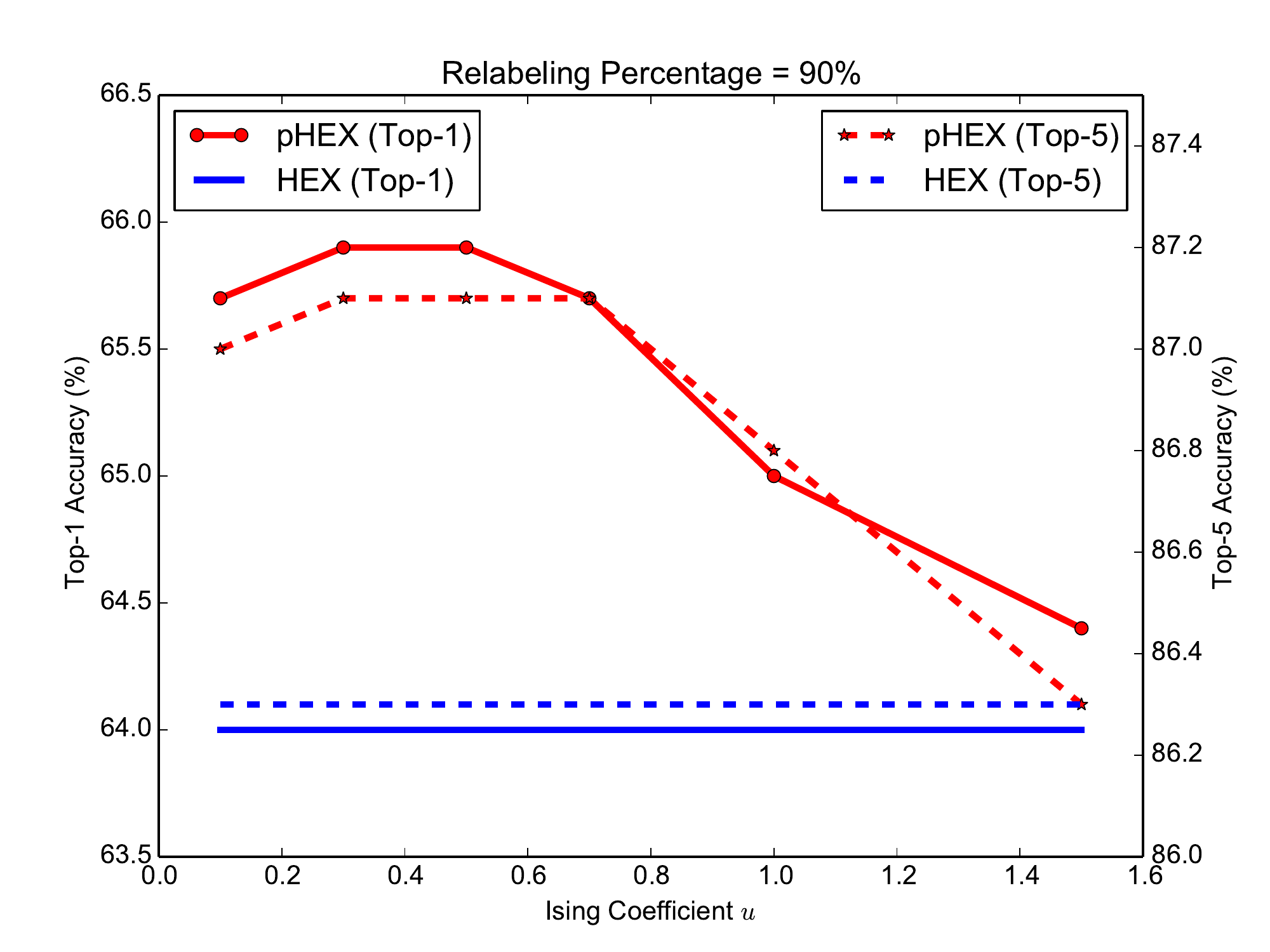}
\includegraphics[width=1.5in]{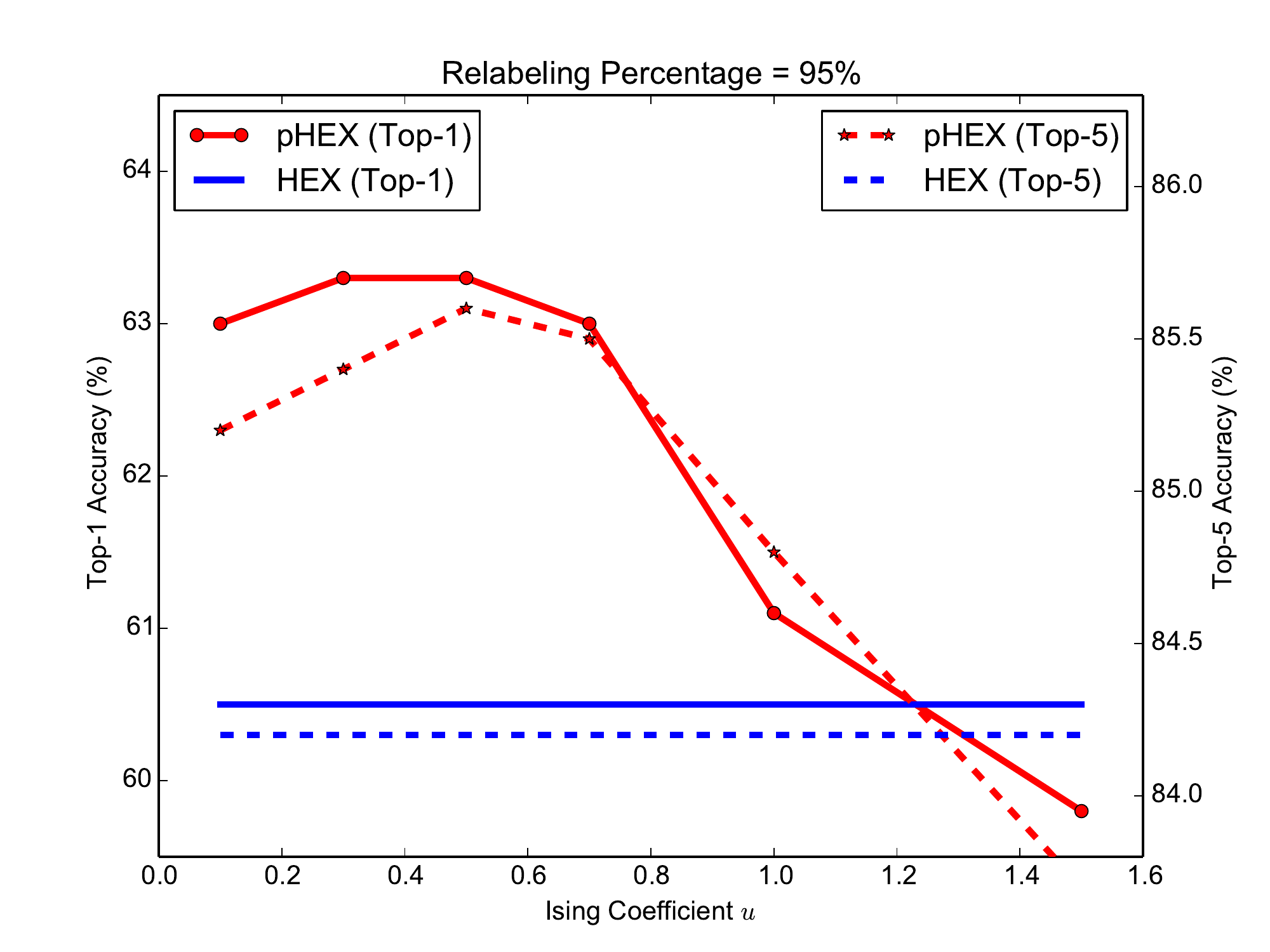}
\includegraphics[width=1.5in]{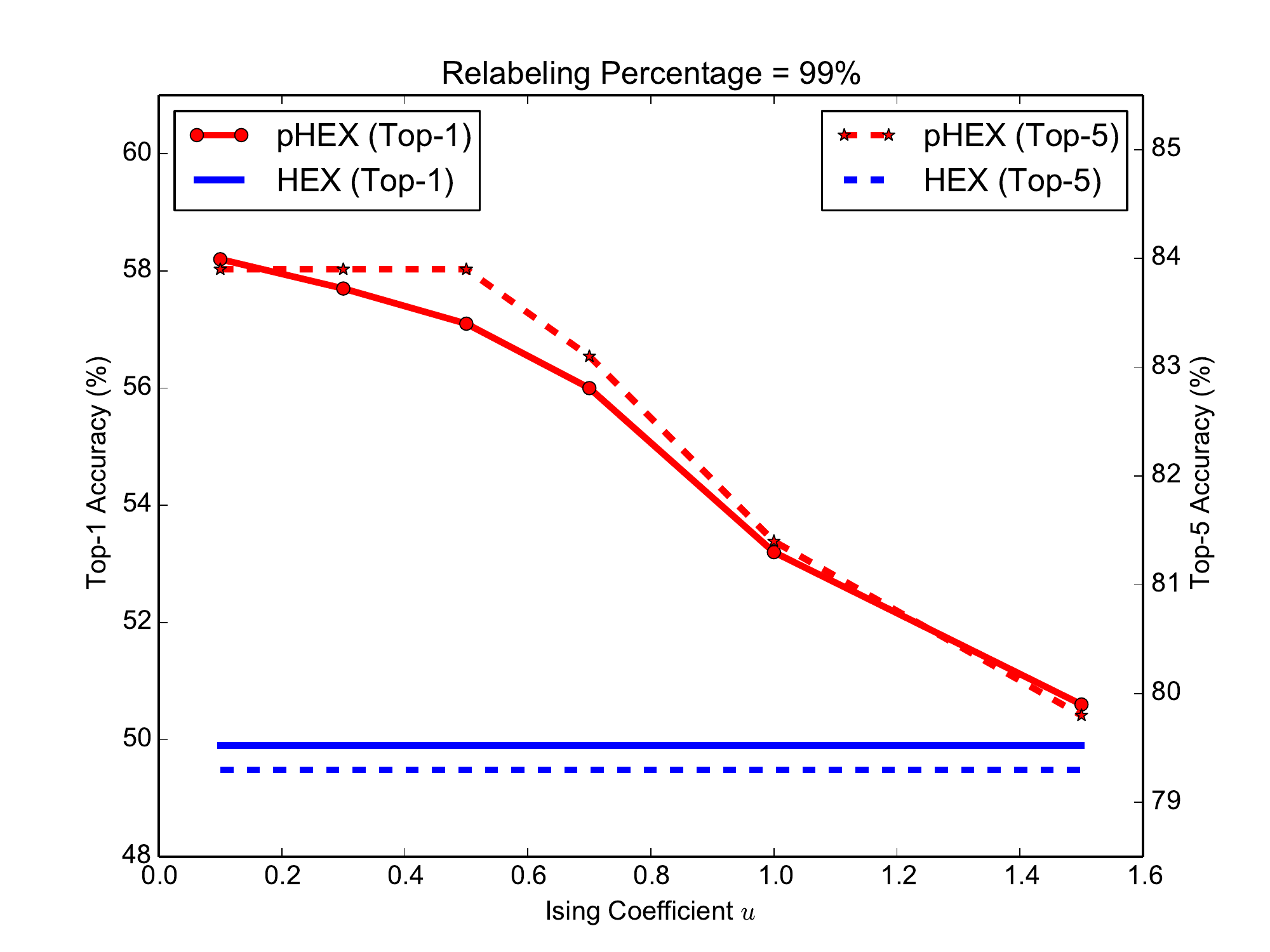}
\caption{Mean Accuracies of the ILSVRC Relabeling Experiments.  }
\label{fig:imagenet}
\end{figure}
}

\eat{
Table \ref{table:imagenet} shows the top 1 and top 5 classification accuracies on the leaf nodes with different amounts of training label relabeling $\cbr{50\%, 90\%, 95\%, 99\%}$ in the ILSVRC2012 dataset. As a reference, the accuracy of all examples labeled at leaf nodes (0\% relabeling) is 72.2\% at top 1. 

When $u$ is smaller, the dependencies between the labels become weaker and the relations are less absolute. We find that the non-absolute label relations provide a better way of modeling the relationship between the labels. Especially, when $u \le 1$, the pHEX graph appears to perform consistently better than the HEX graph in all the relabeled datasets \footnote{Note that although cross validation results are not available in this dataset, it is widely considered that any $\ge 1\%$ top-1 improvement on the ILSVRC2012 dataset is considered to be significant. }. At the optimal weights, the top-1 accuracies of the pHEX graphs improves 0.3\%, 1.9\%, 2.6\%, 8.3\% compared to the accuracies of the HEX graphs for 50\%, 90\%, 95\%, 99\% relabeling respectively. 
}

Figure~\ref{fig:imagenet} shows the Top-1 (top row) and Top-5 (bottom row) accuracies across classes as a function of $u$, for the relabeling experiments. For comparison, the Top-1 (top row) and Top-5 (bottom row) accuracies without relabeling (i.e., the standard ImageNet setup) is 70.1\% and 90.0\% respectively. Not surprisingly, relabeling
(i.e.,  only providing some labels at the leaves, and using coarser grained categories
for the rest) hurts performance (as estimated by leaf-level accuracy). However,
in this regime (which occurs commonly in practice), pHEX generally outperforms HEX, especially
for 90\%, 95\% and 99\% relabeling, where the accuracies improve by 2\%, 3\% and 8\% respectively. 
(Note that a 1\% difference in performance  is considered statistically significant on this problem due to the large size of this dataset.)

At first, it might seem odd that relaxing the hard constraints imposed by the hierarchy can help,
since the hierarchy provided by WordNet is supposed to be correct.
However,
\cite{DenDinJia14} observed that too few training examples labeled at leaf nodes (especially at 99\% relabeling) may confuse  the leaf models, especially at the beginning of the training. As the algorithm runs longer, it becomes harder to recover from a bad local minimum because the constraints in the HEX graph are hard constraints. 
By contrast, in the pHEX graph, the weaker relations between internal nodes and leaf nodes make the resulting posterior distribution smoother, so it is easier to overcome bad local minima for the pHEX graph in later iterations.

It is also interesting to see that the optimal value of $u$ appears to depend on the relabeling percentage. When a larger portion of training examples are relabeled, e.g. 99\% relabeling, the optimal relation coefficient becomes weaker $(u=0.1)$. This indicates that weaker label relations are preferred when there is more uncertainty in the leaf labels. 

On the other hand, when $u$ is large, the label relations become quite certain and the the pHEX graph becomes closer to the HEX graph. In the case of $u = 1.5$ ($q \simeq 0.002$), the performance of pHEX graph can become worse than HEX graph,
probably due to the inability to perform exact inference in the pHEX graph.

\subsection{Zero shot learning experiments}

\begin{figure*}[htp]
\centering
\includegraphics[width=2.2in]{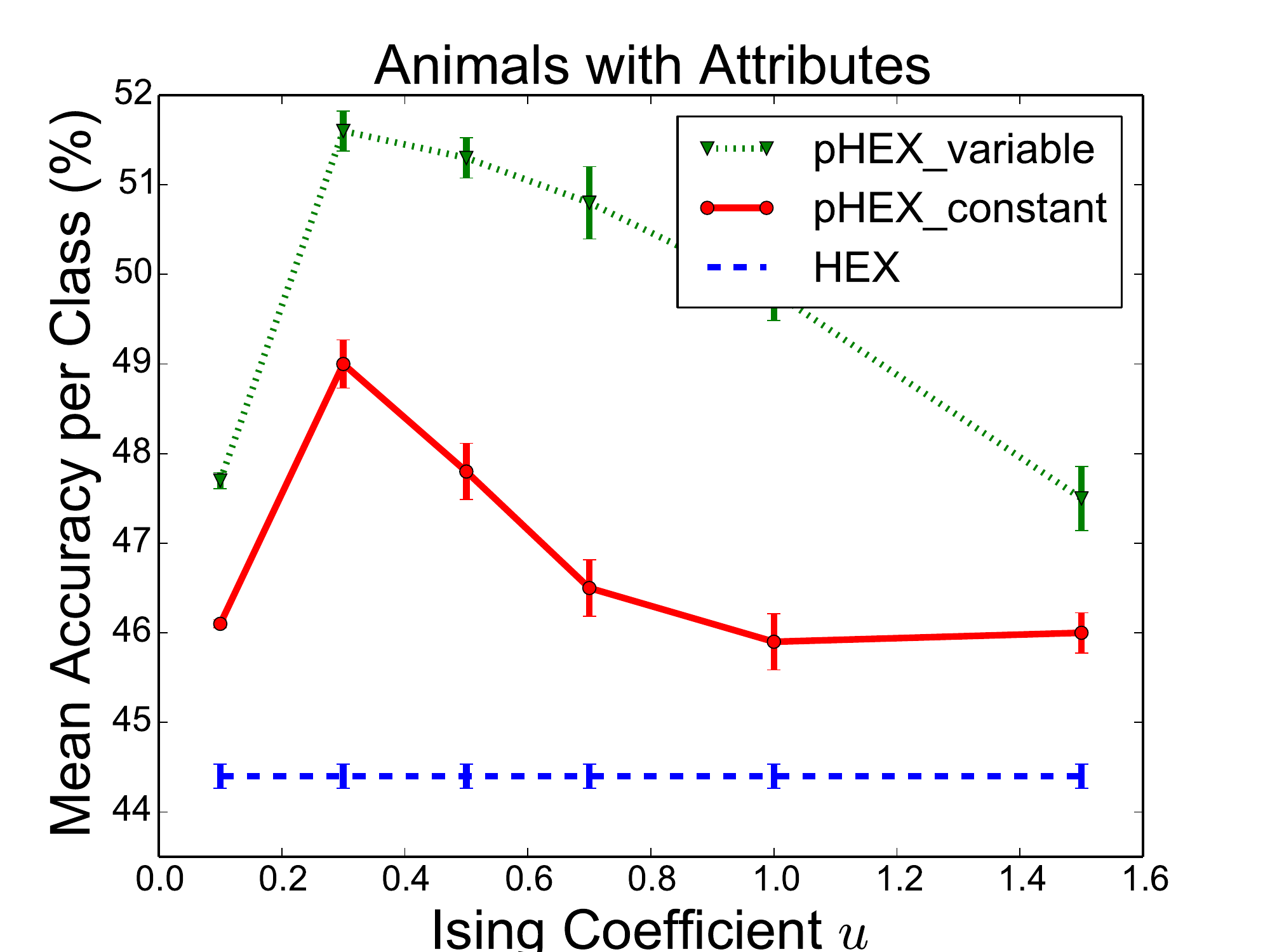}
\includegraphics[width=2.2in]{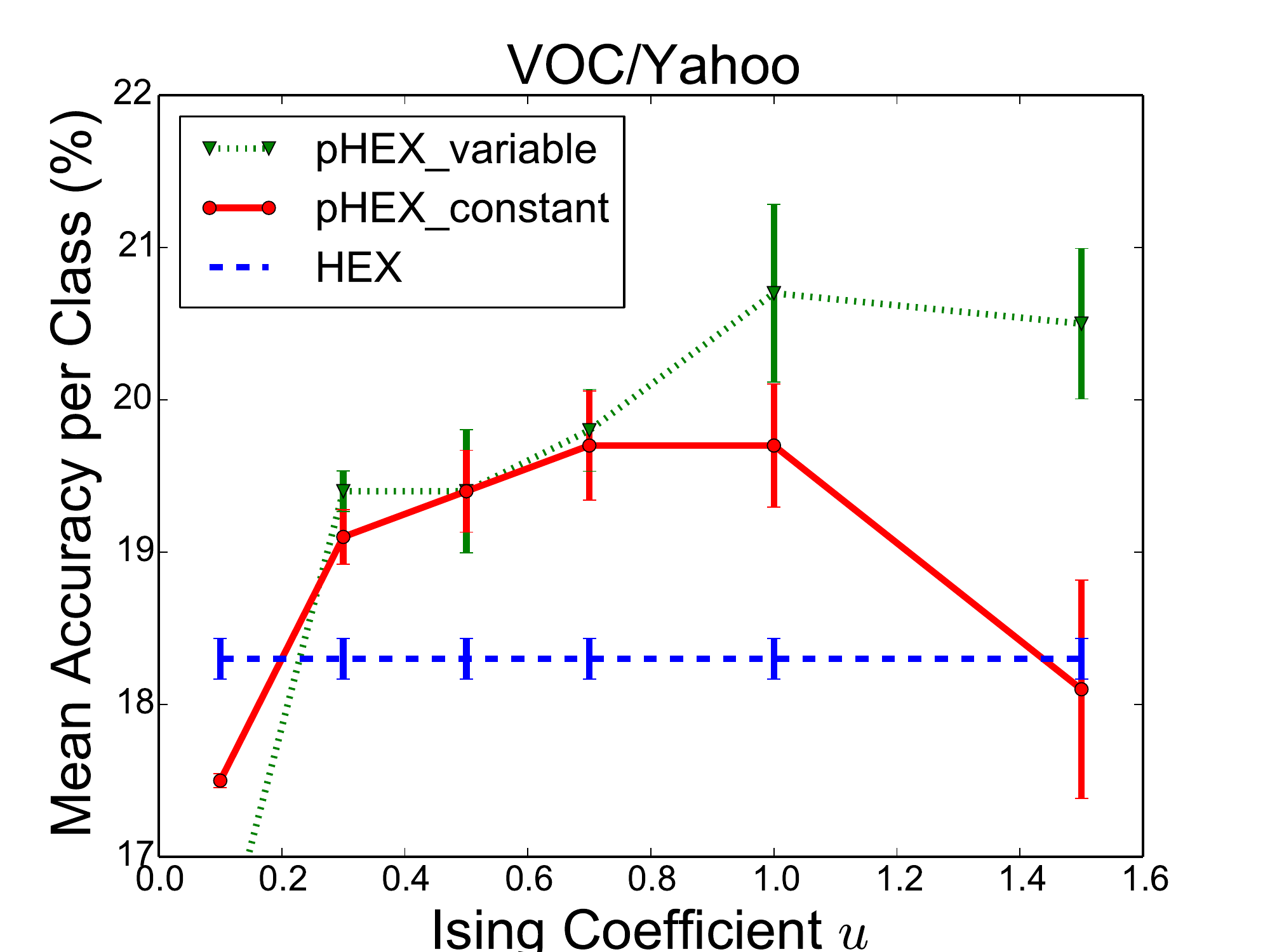}
\caption{Mean accuracy per class vs relation strength $u$ for the Zero-shot Learning Experiments.  
Left: animals with attributes. The results of the pHEX with variable edge weights are in the dotted green,
 and the ones with constant edge weights are in solid red.
The results of the HEX graphs are in the blue dashed horizontal lines. 
Right: VOC/Yahoo images with attributes.
}
\label{fig:zero-shot}
\end{figure*}

We use two datasets to illustrate zero shot learning.
The first is the Animals with Attributes dataset \cite{lampert2009},
which  includes images from 50 animal classes. For each animal class, it provides both binary and continuous predicates for 85 attributes. We convert the binary predicates to constant (soft) relations, and the continuous predicates to variable soft relations by a monotonic mapping function. The details of the mapping are provided in the supplementary material. We evaluate the zero-shot setting where training is performed using only examples from 40 animal classes (with 24295 images) and testing is on classifying the 10 unseen classes (with 6180 images). Our experimental results are based on 5-fold cross validation. The underlying network is a single-layer network whose inputs come from the recently released DECAF features \cite{Donahue2014}.

The second dataset is the
aPascal-aYahoo dataset \cite{Farhadi09},
which consists of a 12695 image subset of the PASCAL VOC 2008 dataset and 2644 images that
were collected using the Yahoo image search engine. The PASCAL part serves as training data and has 20 object classes. The Yahoo part serves as test data and contains 12 different object classes. Each image has been annotated with 64 binary attributes that characterize shape, material and the presence of important parts of the visible object. We convert them to binary and continous predicates for attributes per object by averaging the image annotations for every object (details in supplementary material). The underlying network is again a single-layer network whose inputs come from the features that the authors of \cite{Farhadi09} extracted from the objects bounding boxes (as provided by the PASCAL VOC annotation) and released as part of the dataset. Once again we use 5-fold cross validation and compares constant soft relations and variable soft relations with hard relations.

\eat{
Table \ref{table:zero-shot} shows the mean accuracy per class, standard deviation, as well as the p-values of a paired t-test between the results of the pHEX graphs and the HEX graphs on the two zero-shot learning tasks: Animals with Attributes and VOC-Yahoo. 
Similar to the ILSVRC2012 dataset, in the two zero-shot learning tasks we find that the pHEX graph improves the mean accuracy over the HEX graph with certain Ising coefficients $u$. In the Animal and Attirbutes dataset, the pHEX graph achieves the optimal relation weights at $u = 0.3$, and improve the mean accuracies over the HEX graph by 4.6\%. In the VOC-Yahoo dataset, the pHEX graph achieves optimal weights at $u=0.7$ or $1.0$, and improves 1.4\% accuracy over the HEX graph. Both improvements are significant according to the p-values ($\le 5\%$) of the paired t-test: the p-value is 0.0\% for Animals with Attributes and 0.8\% for VOC-Yahoo.  The accuracies of the pHEX graph gets closer to the ones of the HEX graph as $u$ becomes larger and the pHEX graph approaches to the HEX graph. 
}

Figure~\ref{fig:zero-shot} shows the mean accuracy per class (along with standard errors) vs $u$.
We see that pHEX is generally significantly outperforming HEX.
In particular, when $u \in [0.1,1.5]$ for Animals with Attributes and $u \in [0.3,1.0]$ for VOC/Yahoo, the difference is statistically significant at the 5\%  level according to a paired t-test.
 The accuracies of the pHEX graph get closer to the ones of the HEX graph as $u$ becomes larger and the pHEX graph approaches to the HEX graph. Moreover, the pHEX models with variable soft relations improves over the ones with constant soft relations by 2\% for Animals with Attributes and 1\% for VOC/Yahoo. This demonstrates the value of adding additional information in the variable probabilistic label relations in transfer learning.

\eat{
\begin{figure}[htp]
\centering
\includegraphics[width=1.5in]{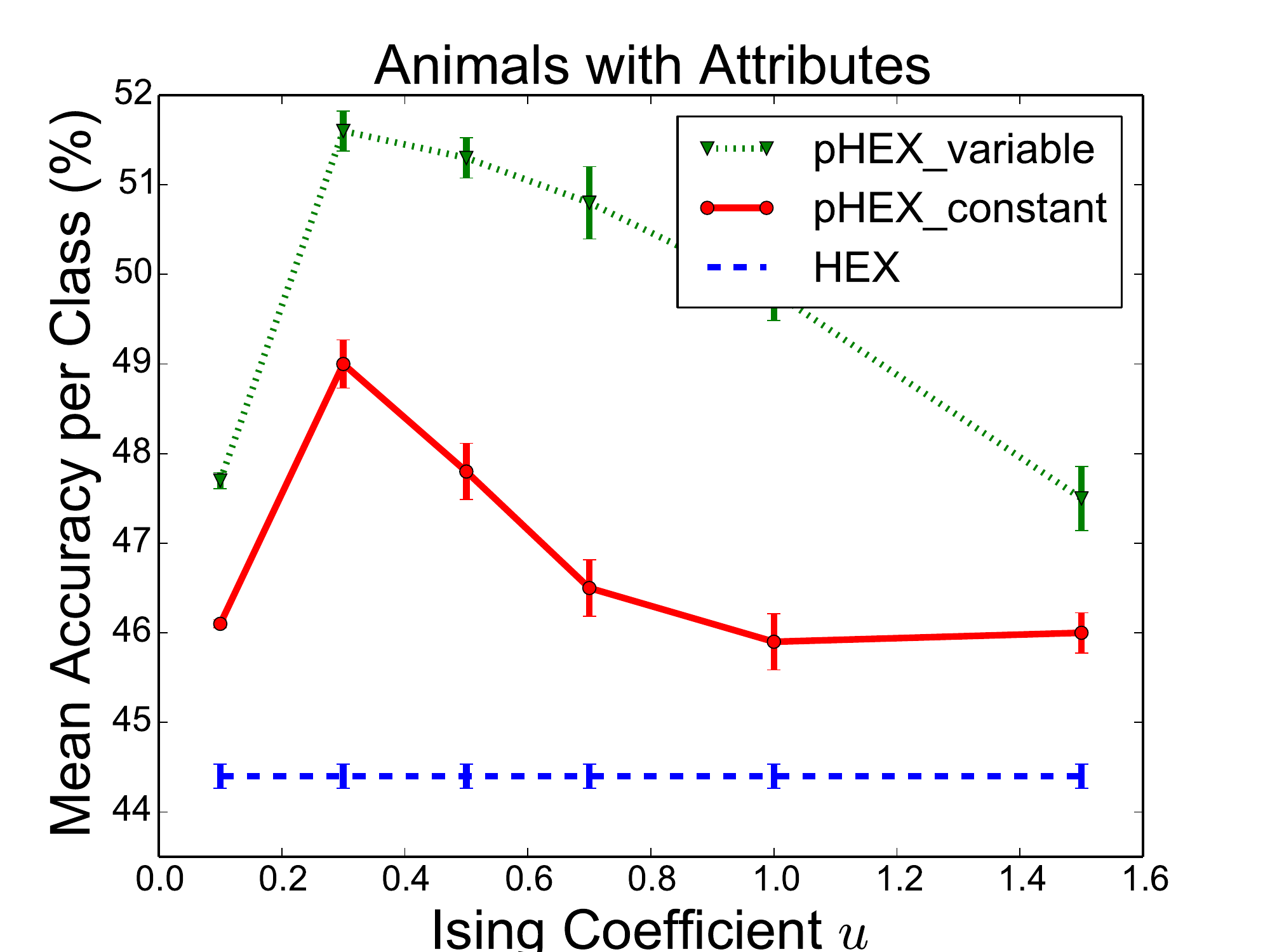}
\includegraphics[width=1.5in]{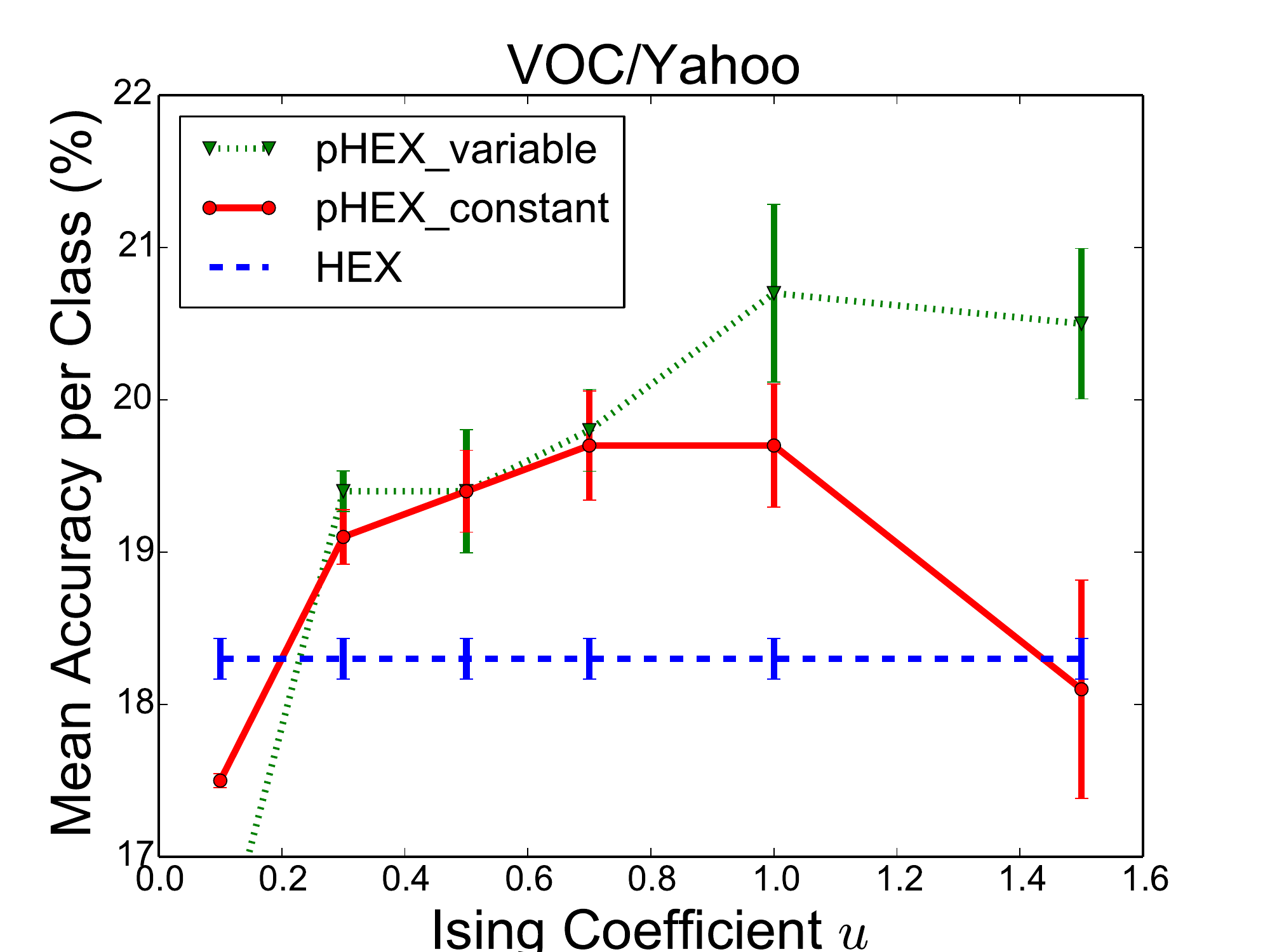}
\caption{Mean Accuracies of the Zero-shot Learning Experiments.  }
\label{fig:zero-shot}
\end{figure}
}

\eat{
\begin{table}[htp]
  \caption{Mean accuracy per class $\pm$ standard deviation (the p-values as compared to the HEX graph in brackets) on the two zero-shot learning datasets.}
  \begin{center}
    \begin{tabular}{||c||c|c||}
      \hline
      Dataset & Animals with Attributes & VOC-Yahoo\\
      \hline
	$u=0.1$ & 46.1 $\pm$ 0.1 (0.0\%) & 17.5 $\pm$ 0.1 (0.1\%) \\
	$u=0.3$ & {\bf 49.0 $\pm$ 0.6 (0.0\%)} & 19.1 $\pm$ 0.4 (4.2\%)\\
	$u=0.5$ & 47.8 $\pm$ 0.7 (0.1\%) &  19.4 $\pm$ 0.6 (0.2\%)\\
	$u=0.7$ & 46.5 $\pm$ 0.7 (0.3\%)  & {\bf 19.7 $\pm$ 0.8 (0.8\%)}\\
	$u=1.0$ & 45.9 $\pm$ 0.7 (1.3\%)  & {\bf 19.7 $\pm$ 0.9 (0.5\%)}\\
	$u=1.5$ & 46.0 $\pm$ 0.5 (0.2\%) & 18.1 $\pm$ 1.6 (81.8\%)\\
      \hline
      HEX & 44.4 $\pm$ 0.3 & 18.3 $\pm$ 0.3 \\
	\hline
    \end{tabular}
  \end{center}
  \label{table:zero-shot}
\end{table}
}

\subsection{Speed comparison of HEX vs pHEX}

\eat{
The inference cost of the pHEX graph is proportional to the number of probabilistic relations (edges) in the graph. In the 
ImageNet experiment, the number of edges is about 6000, which is about
3 times larger than the number of nodes in graph
(which is 1860). However, the total cost of inference is negligible compared to the cost of the underlying
deep neural network.
}

In the ImageNet experiements, the cost of HEX and pHEX is similar,
since most of the time is spent evaluating the underlying deep CNN.
In the two zero-shot learning experiments, the inference time of the pHEX graph is about the same as the one of the HEX graph.
Furthermore, many other algorithms such as quantum annealing \cite{KadNis98} (which are faster and/or more
accurate than loopy belief propagation) have been devised for Ising models which we could try in the future.

\section{Conclusions}
\label{sec:conclusion}
In this paper, we studied object classification with probabilistic label relations.
In particular, we proposed the pHEX graph, which naturally generalizes the HEX graph. The pHEX graph is equivalent to an undirected Ising model, which allows for efficient approximate inference methods. We embed the pHEX graph 
on top of a deep neural network,
and show that it outperforms the HEX graph on a number of classification tasks which require exploiting label relations. 

There are several possible future directions of this work. One idea
is to learn the Ising coefficients of the pHEX graph together with the underlying neural network parameters. Another
is to combine the pHEX graph into a larger framework which exploits spatial relations between  objects.


%
\bibliographystyle{ieee}
\bibliography{phex}  

\begin{thebibliography}{10}\itemsep=-1pt

\bibitem{AkataPHS13}
Z.~Akata, F.~Perronnin, Z.~Harchaoui, and C.~Schmid.
\newblock Label-embedding for attribute-based classification.
\newblock In {\em CVPR}, pages 819--826. IEEE, 2013.

\bibitem{BenLouCol2009}
Y.~Bengio, J.~Louradour, R.~Collobert, and J.~Weston.
\newblock Curriculum learning.
\newblock In {\em ICML}, 2009.

\bibitem{Binder01}
K.~Binder.
\newblock Ising model, 2001.
\newblock Encyclopedia of Mathematics, Springer, ISBN 978-1-55608-010-4.

\bibitem{Bishop2006}
C.~M. Bishop.
\newblock {\em Pattern Recognition and Machine Learning}.
\newblock Springer-Verlag New York, Inc., Secaucus, NJ, USA, 2006.

\bibitem{chen2011multi}
X.~Chen, X.-T. Yuan, Q.~Chen, S.~Yan, and T.-S. Chua.
\newblock Multi-label visual classification with label exclusive context.
\newblock In {\em ICCV}, pages 834--841. IEEE, 2011.

\bibitem{Dalvi2015}
B.~Dalvi, E.~Minkov, P.~P. Talukdar, and W.~W. Cohen.
\newblock Automatic gloss finding for a knowledge base using ontological
  constraints.
\newblock In {\em WSDM}, WSDM '15, pages 369--378. ACM, 2015.

\bibitem{deng2012}
J.~Deng, A.~Berg, S.~Satheesh, H.~Su, A.~Khosla, and L.~Fei-Fei.
\newblock Imagenet large scale visual recognition challenge 2012, 2012.
\newblock www.image-net.org/challenges/LSVRC/2012.

\bibitem{DenDinJia14}
J.~Deng, N.~Ding, Y.~Jia, A.~Frome, K.~Murphy, S.~Bengio, Y.~Li, H.~Neven, and
  H.~Adam.
\newblock Large-scale object classfication using label relation graphs.
\newblock In {\em ECCV}, pages 48--64, Zurich, Switzerland, September 2014.
  Springer.

\bibitem{Donahue2014}
J.~Donahue, Y.~Jia, O.~Vinyals, J.~Hoffman, N.~Zhang, E.~Tzeng, and T.~Darrell.
\newblock Decaf: A deep convolutional activation feature for generic visual
  recognition.
\newblock In {\em ICML}, 2014.

\bibitem{Farhadi09}
A.~Farhadi, I.~Endres, D.~Hoiem, and D.~Forsyth.
\newblock Describing objects by their attributes.
\newblock In {\em CVPR}, 2009.

\bibitem{KadNis98}
T.~Kadowaki and H.~Nishimori.
\newblock Quantum annealing in the transverse ising model.
\newblock {\em Physics Review E}, 58, 1998.

\bibitem{Lafferty01crf}
J.~D. Lafferty, A.~McCallum, and F.~C.~N. Pereira.
\newblock Conditional random fields: Probabilistic models for segmenting and
  labeling sequence data.
\newblock In {\em ICML}, ICML '01, pages 282--289, San Francisco, CA, USA,
  2001.

\bibitem{lampert2009}
C.~H. Lampert, H.~Nickisch, and S.~Harmeling.
\newblock Learning to detect unseen object classes by between-class attribute
  transfer.
\newblock In {\em CVPR}, pages 951--958. IEEE, 2009.

\bibitem{marszalek2007}
M.~Marszalek and C.~Schmid.
\newblock Semantic hierarchies for visual object recognition.
\newblock In {\em CVPR}, pages 1--7. IEEE, 2007.

\bibitem{MirRavXu2015}
F.~Mirzazadeh, S.~Ravanbakhsh, B.~Xu, N.~Ding, and D.~Schuurmans.
\newblock Embedding inference for structured multilabel prediction.
\newblock In {\em NIPS}, 2015.

\bibitem{Ordonez2013}
V.~Ordonez, J.~Deng, Y.~Choi, A.~C. Berg, and T.~L. Berg.
\newblock From large scale image categorization to entry-level categories.
\newblock In {\em ICCV}, 2013.

\bibitem{Palatucci09}
M.~Palatucci, G.~Hinton, D.~Pomerleau, and T.~M. Mitchell.
\newblock Zero-shot learning with semantic output codes.
\newblock In {\em NIPS}, 2009.

\bibitem{Richardson06}
M.~Richardson and P.~Domingos.
\newblock Markov logic networks.
\newblock {\em Machine Learning}, 62:107--136, 2006.

\bibitem{rohrbach11cvpr}
M.~Rohrbach, M.~Stark, and B.~Schiele.
\newblock Evaluating knowledge transfer and zero-shot learning in a large-scale
  setting.
\newblock In {\em CVPR}, 2011.

\bibitem{RohrbachSSGS10}
M.~Rohrbach, M.~Stark, G.~Szarvas, I.~Gurevych, and B.~Schiele.
\newblock What helps where - and why? semantic relatedness for knowledge
  transfer.
\newblock In {\em CVPR}, pages 910--917. IEEE, 2010.

\bibitem{SharmanskaQL12}
V.~Sharmanska, N.~Quadrianto, and C.~H. Lampert.
\newblock Augmented attribute representations.
\newblock In {\em ECCV}, pages 242--255, 2012.

\bibitem{Szegedy2014}
C.~Szegedy, W.~Liu, Y.~Jia, P.~Sermanet, S.~Reed, D.~Anguelov, D.~Erhan,
  V.~Vanhoucke, and A.~Rabinovich.
\newblock Going deeper with convolutions, 2014.
\newblock arXiv:1409.4842.

\bibitem{Tso05svmstruct}
I.~Tsochantaridis, T.~Joachims, T.~Hofmann, and Y.~Altun.
\newblock Large margin methods for structured and interdependent output
  variables.
\newblock {\em J. Mach. Learn. Res.}, 6:1453--1484, Dec. 2005.

\bibitem{Vishy06}
S.~V.~N. Vishwanathan, N.~N. Schraudolph, M.~W. Schmidt, and K.~P. Murphy.
\newblock Accelerated training of conditional random fields with stochastic
  gradient methods.
\newblock In {\em ICML}, ICML '06, pages 969--976, New York, NY, USA, 2006.
  ACM.

\bibitem{wu-RSS-14}
C.~Wu, I.~Lenz, and A.~Saxena.
\newblock Hierarchical semantic labeling for task-relevant rgb-d perception.
\newblock In {\em Proceedings of Robotics: Science and Systems}, Berkeley, USA,
  July 2014.

\bibitem{CVPR13Yu}
F.~Yu, L.~Cao, R.~Feris, J.~Smith, and S.-F. Chang.
\newblock Designing category-level attributes for discriminative visual
  recognition.
\newblock In {\em CVPR}, Portland, OR, June 2013.

\bibitem{zweig2007}
A.~Zweig and D.~Weinshall.
\newblock Exploiting object hierarchy: Combining models from different category
  levels.
\newblock In {\em ICCV}, pages 1--8. IEEE, 2007.

\end{thebibliography}

\end{document}